\documentclass{article}
\usepackage{spconf,amsmath,graphicx}

\title{Dense Optical Flow based Change Detection Network\\ Robust to Difference of Camera Viewpoints}

\name{Ken Sakurada$^{1,2}$ \ \  Weimin Wang$^{1}$ \ \  Nobuo Kawaguchi$^{1}$ \ \ Ryosuke Nakamura $^{2}$ \sthanks{This paper is based on results obtained from a project commissioned by the New Energy and Industrial Technology Development Organization (NEDO).}}
\address{$^1$ Nagoya University, Nagoya, Japan \\$^2$National Institute of Advanced Industrial Science and Technology, Tokyo, Japan}

\begin{document}
\maketitle
\begin{abstract}
This paper presents a novel method for detecting scene changes from a pair of images with a difference of camera viewpoints using a dense optical flow based change detection network.
In the case that camera poses of input images are fixed or known, such as with surveillance and satellite cameras, the pixel correspondence between the images captured at different times can be known. 
Hence, it is possible to comparatively accurately detect scene changes between the images by modeling the appearance of the scene.
On the other hand, in case of cameras mounted on a moving object, such as ground and aerial vehicles, we must consider the spatial correspondence between the images captured at different times. 
However, it can be difficult to accurately estimate the camera pose or 3D model of a scene, owing to the scene changes or lack of imagery.
To solve this problem, we propose a change detection convolutional neural network utilizing dense optical flow between input images to improve the robustness to the difference between camera viewpoints. 
Our evaluation based on the panoramic change detection dataset shows that the proposed method outperforms state-of-the-art change detection algorithms.
\end{abstract}

\begin{figure}[t]
\centering
\includegraphics[width=65mm,bb=0 0 1024 224]{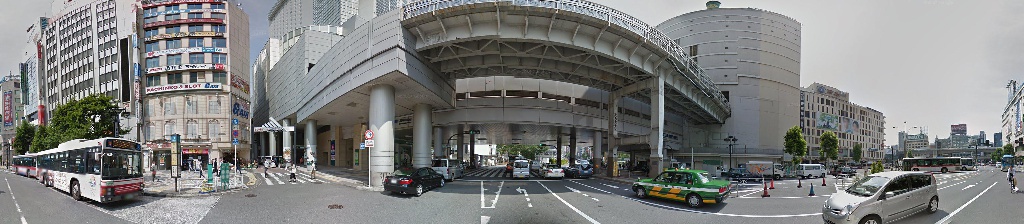}  \\
\vspace{1mm}
\includegraphics[width=65mm,bb=0 0 1024 224]{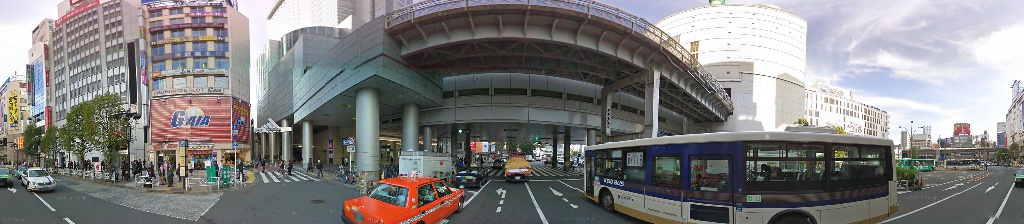} \\
\vspace{1mm}
\includegraphics[width=65mm,bb=0 0 1024 224]{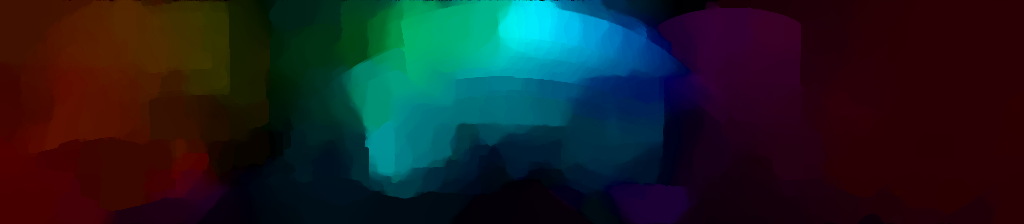} \\
\vspace{1mm}
\includegraphics[width=65mm,bb=0 0 1024 224]{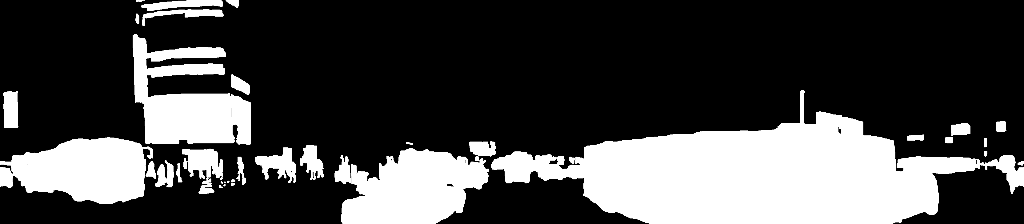}\\
\vspace{1mm}
\includegraphics[width=65mm,bb=0 0 1024 224]{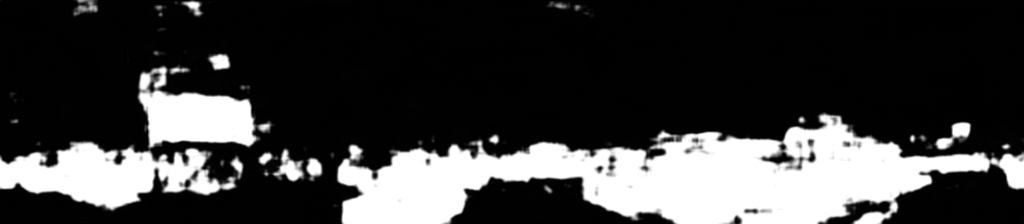} 
\caption{Example of scene change detection using CNN based on dense optical flow. From top to bottom, input image taken at times $t_0$ and $t_1$, the estimated optical flow, the hand-labeled ground-truth and the estimation result of change detection.}
\label{fig:example_scene_change}
\end{figure}

\section{Introduction}
This paper addresses the problem of detecting temporal scene changes from a pair of images captured at two different time points. 
In most change detection problems, it is assumed that input images are accurately registered, especially for surveillance and satellite camera systems \cite{lu2004change,Radke2005,Pollard2007,fujita2017damage}.
However, for moving cameras, such as vehicle-mounted cameras and  mobile devices, we should consider the difference of the camera poses of the input images, because it is difficult to capture a scene from similar viewpoints every time due to the high flexibilities of camera pose and shutter timing.

Although structure-from-motion (SfM) \cite{Agarwal2009,frahm2010building} is utilized in many change detection methods to estimate the camera poses of input images, the estimated camera poses can be unreliable due to the lack of the feature correspondence between the images due to the scene changes.
Furthermore, for city-scale problems, computational costs of pixel-level image registration via batch optimization (e.g., bundle adjustment) \cite{Triggs1999}, and dense 3D reconstruction based on multi-view stereo \cite{Seitz2006,Furukawa2010,jancosek2011multi} are prohibitively high.
Thus, it is preferable to detect scene changes from roughly registered images using embedded location information, such as Global Positioning System (GPS) metadata.

Several methods based on deep neural networks have been proposed for scene change detection thus far \cite{BMVC2015_127,alcantarilla2016street,khan2017learning}.
However, they assumed that pixel correspondences, or relative camera poses between input images, are known.
The work by \cite{sakurada2015change} proposed a change detection method that differentiates the feature maps extracted from input images utilizing convolutional neural networks (CNNs) \cite{vgg_SimonyanZ14a} trained with ImageNet \cite{russakovsky2015imagenet} or SUN \cite{xiao2010sun} dataset.
Although an advantage of this method is the high generalizability of the trained networks (i.e. scene-independence) owing to the large-scale image dataset, they are not optimized for change detection.

This paper presents a novel method for detecting scene changes from a roughly registered image pair using a dense optical flow based change detection network (DOF-CDNet).
To the best of our knowledge, the proposed networks are the first CNNs considering the difference of camera viewpoints of an input image pair for the kinds of change detection tasks defined in this work.
Specifically, we improve the robustness to the difference of camera viewpoints by the adding optical flow information between the image pairs as the input to the proposed change detection network (Figure \ref{fig:example_scene_change}).

\section{Related Work}
In the fields of computer vision and remote sensing, there has been the extensive study of change detection between images captured at different times, such as scene anomaly detection from surveillance camera images and urbanization and deforestation monitoring by detecting land surface changes from satellite images.
In recent years, several methods of ground-level, wide area scene change detection from images captured by vehicle-mounted cameras and mobile devices have been proposed for the purposes of updating 3D maps for autonomous driving, infrastructure inspection, disaster response, and agricultural automation \cite{BMVC2015_127,sakurada2014massive}.

In the case of surveillance and satellite cameras, the pixel correspondences between images taken at different time points are almost always known.
The work by \cite{mukojima2016moving} proposed a background subtraction method for obstacle detection on railway tracks.
Although the camera mounted on a railway train moves, it always follows the same path and the difference between the camera viewpoints dependent on the shutter timing is small.  
Hence, the scene changes can be reasonably detected with accuracy by modeling the appearance of the scene \cite{radke2005image}.

On the other hand, in the case of cameras mounted on moving objects such as cars and mobile robots, it is necessary to consider spatial correspondences between images captured at different time points. 
If there are enough common parts that are useful as visual feature points between the scenes at different time points, it is possible to estimate relative camera pose utilizing SfM and detect structural scene changes based on multi-view geometry \cite{Taneja2013,Sakurada2013}.
Furthermore, deconvolutional networks for scene change detection has been proposed as a technology that learns the change mask from input RGB images which are aligned by their estimated depths \cite{alcantarilla2016street}.
\begin{figure*}[!t]
  \hfil\includegraphics[width=14cm,bb=100 0 850 150]{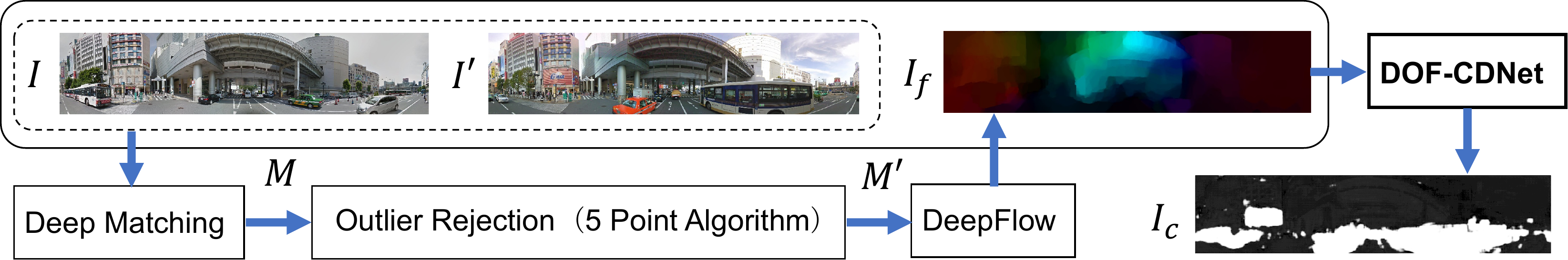}
\vspace{0mm}
  \caption{Flowchart of the proposed change detection method}
  \label{fig:flowchart}
\end{figure*}
However, in the cases of a drastic scene change or an insufficient number of images, it is difficult to accurately estimate camera pose owing to lack of common feature points between images. 
Additionally, the computation cost of city-scale SfM is prohibitively expensive.

In this study, scene changes are detected from an image pair roughly aligned with GPS data. 
So far, for the same purpose, a method has been proposed that extracts grid features utilizing CNNs trained with large-scale image classification datasets, such as ImageNet \cite{russakovsky2015imagenet} and SUN \cite{xiao2010sun}, and detects the scene changes differentiating the grid features \cite{sakurada2015change}．
The method has high generalization capability and less dependence on a target scene since the network is trained with a large-scale image dataset.
However, it also has a problem that the network is not optimized for scene change detection.

Although several convolutional networks for change detection have been proposed \cite{BMVC2015_127,khan2017learning}, they require the additional information of pixel-level correspondences between input images, such as 3D model and depth imagery of a scene.
Therefore, we propose a novel change detection method for the robustness to the difference of camera viewpoints, for which the CNN is trained by inputting not only an RGB image pair, but also the estimated dense optical flow (Figure \ref{fig:example_scene_change}).

\section{Scene Change Detection Dataset}
There are several publicly available change detection datasets acquired by satellite, surveillance and vehicle-mounted cameras \cite{wang2014cdnet}.
In the most of these datasets, either the pixel-level correspondence between input images is known, or it is possible to densely reconstruct the scenes from a sufficient number of multi-view images.

However, in case of a city or regional scale modeling from ground-level images, capturing an entire city with surveillance cameras is infeasible.
Additionally, to detect structural scene changes of an entire city from movies captured by vehicle-mounted cameras based on multi-view geometry, a large amount of image data and computational resources are necessary \cite{Taneja2013,Sakurada2013}.
For applications such as autonomous driving and pedestrian navigation that need frequent and sequential updating 3D maps, it is necessary to monitor whole-city changes with low-cost change detection methods and then to accurately remeasure the change areas using autonomous agents, such as self-driving cars.

The objective of this study is detecting scene changes from an image pair roughly aligned with GPS information, instead of that is accurately aligned with methods like relative pose estimation, to reduce the computational costs.
Thus, we evaluate our method on the panoramic change detection (PCD) dataset \cite{sakurada2015change}, in which image pairs are roughly aligned.
This dataset consists of two subsets, named ``TSUNAMI" and ``GSV," and each subset consists of 100 panoramic image pairs and the hand-labeled change mask of each pair.
The camera viewpoints of each image pair are different, because the images are captured by a vehicle-mounted camera every few months or years.  
Therefore, it is necessary to detect scene changes based on the difference between camera viewpoints.

\begin{figure}[t]
\begin{center}
\includegraphics[width=65mm,bb=0 0 1130 517]{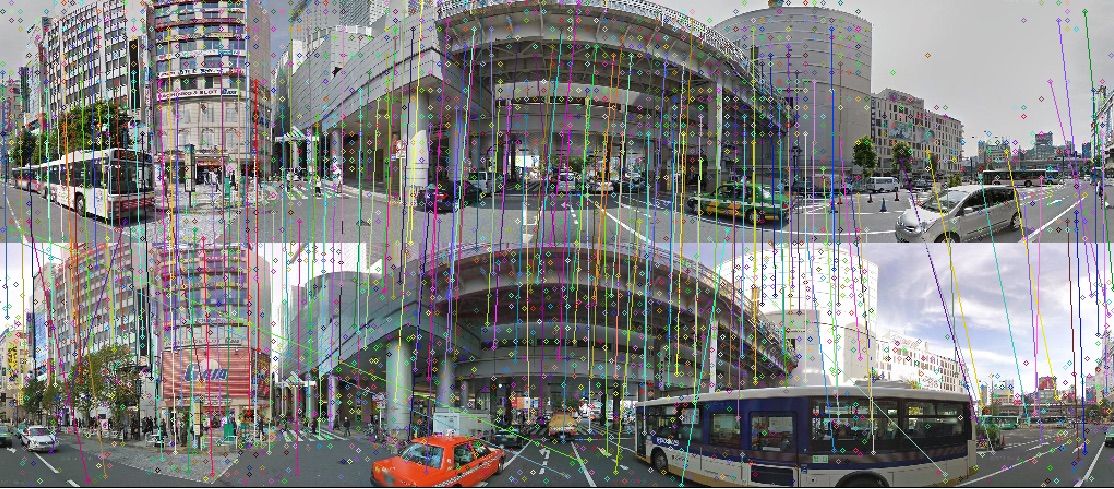}  \\
\includegraphics[width=65mm,bb=0 0 1130 517]{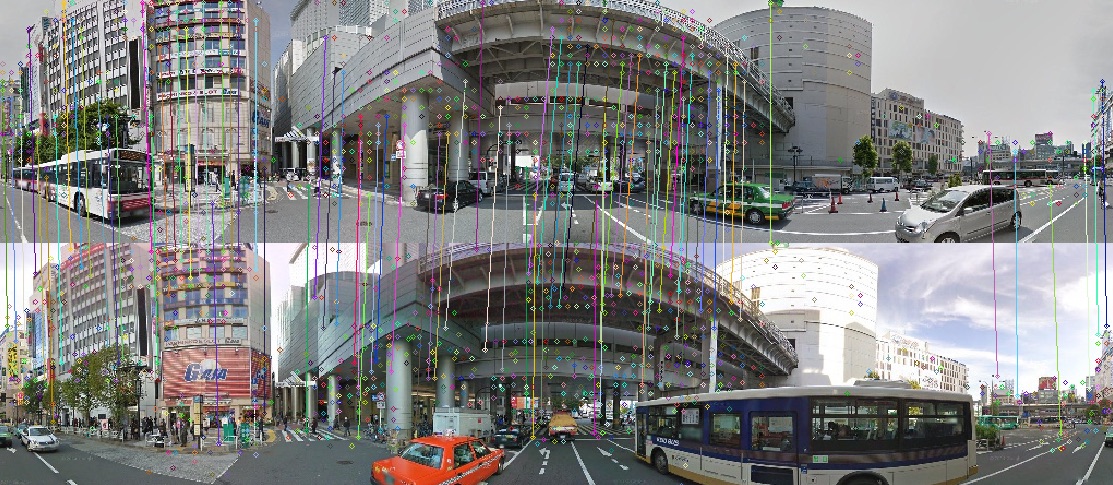} 
\caption{Example of feature matching based on deep matching before (top) and after (bottom) outlier rejection by RANSAC whose model is the five-point algorithm.}
\label{fig:deepmatch}
\vspace{0mm}
\end{center}
\end{figure}

\section{Dense Optical Flow based ConvNet}
To improve robustness against the difference of camera viewpoints, we propose DOF-CDNet, which estimates scene change probability of each pixel between input images utilizing not only the RGB images but also the estimated dense optical flow.
Figure \ref{fig:flowchart} shows the flowchart of the proposed method.
The details of the dense optical flow estimation and the network architecture are described below.

\subsection{Dense Optical Flow Estimation}
\label{subsec:optical_flow}
There are various types of dense optical flow estimation methods based on image features, learning algorithms, etc. \cite{dosovitskiy2015flownet}.
The proposed method of this study exploits DeepFlow \cite{weinzaepfel2013deepflow}, which is not based on learning algorithms, but is extended to add geometric constrains (Figure \ref{fig:flowchart}).

More concretely, tentative matching points between input images $I$, $I'$ captured at time $t_0$, $t_1$ are calculated by DeepMatching.
From these tentative matching points, outliers are removed by random sample consensus with the model defined by the five-point algorithm \cite{nister2004efficient} (Figure \ref{fig:deepmatch}).
The optical flow of each pixel is estimated using only the inliers. 
The outlier removal, which uses epipolar constraint, can improve the estimation accuracy of optical flow, because there are scene changes without correspondence between input images.

\begin{table}[tb]
\vspace{-3mm}
\begin{center}
\caption{Network architecture of DOF-CDNet \label{tab:CNN_Structure}}{
\scalebox{1.0}[1.0]{
\begin{tabular}{c|c}
\hline
Encoder & Decoder \\ \hline
CR~(64, 3, 1)  & CBRD~(512, 4, 2)  \\
CBR~(128, 4, 2) & CBRD~(512, 4, 2) \\ 
CBR~(256, 4, 2) & CBRD~(512, 4, 2) \\ 
CBR~(512, 4, 2) & CBR~(512, 4, 2) \\
CBR~(512, 4, 2) & CBR~(256, 4, 2) \\
CBR~(512, 4, 2) & CBR~(128, 4, 2)  \\
CBR~(512, 4, 2) & CBR~(64, 4, 2)  \\
CBR~(512, 4, 2) & C~(1, 3, 1)  \\ \hline
\end{tabular} 
}
}
\end{center}
\end{table}

\begin{figure*}[!t]
\begin{center}
\vspace{0mm}
$\begin{array}{p{55mm}p{55mm}p{55mm}}
\hspace*{0mm}\includegraphics[width=55mm,bb=0 30 377 239]{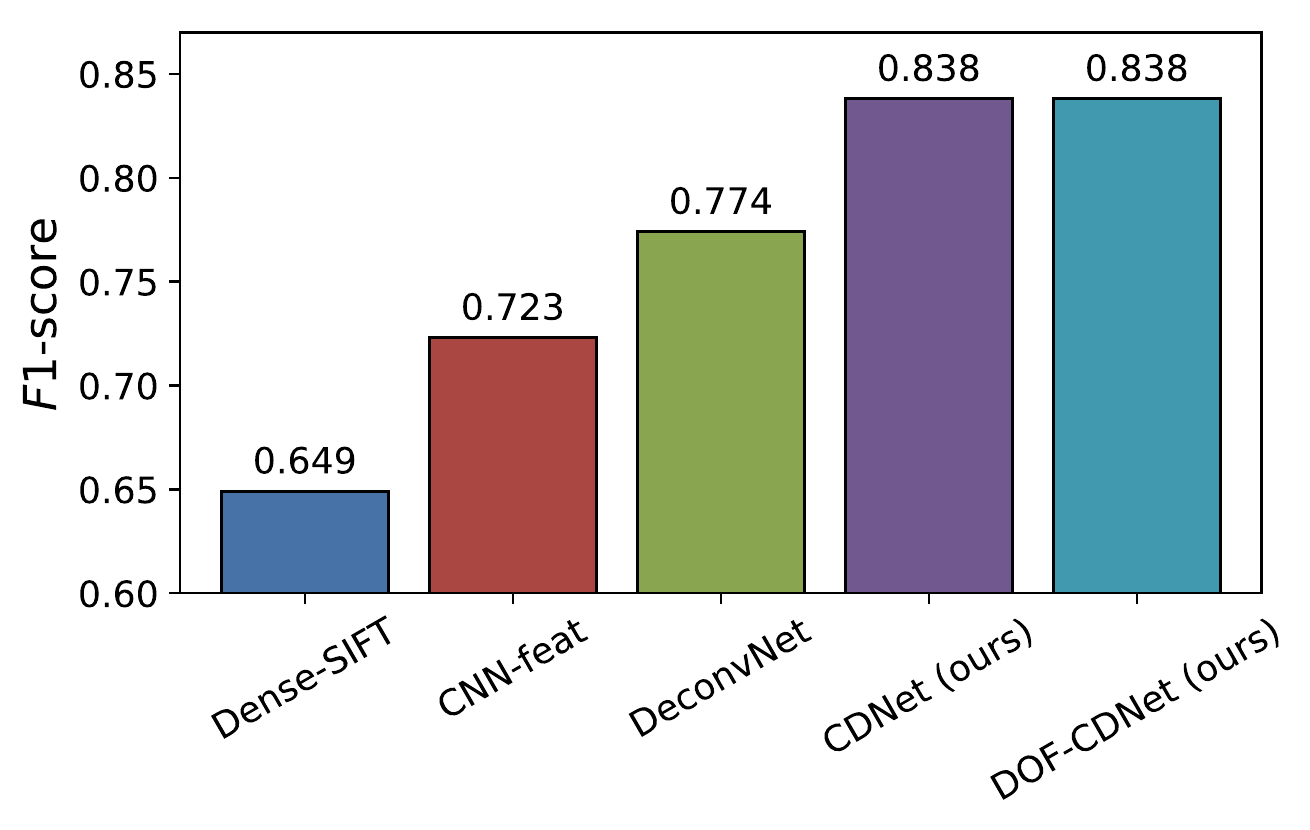} 
&\hspace*{0mm}\includegraphics[width=55mm,bb=0 30 377 239]{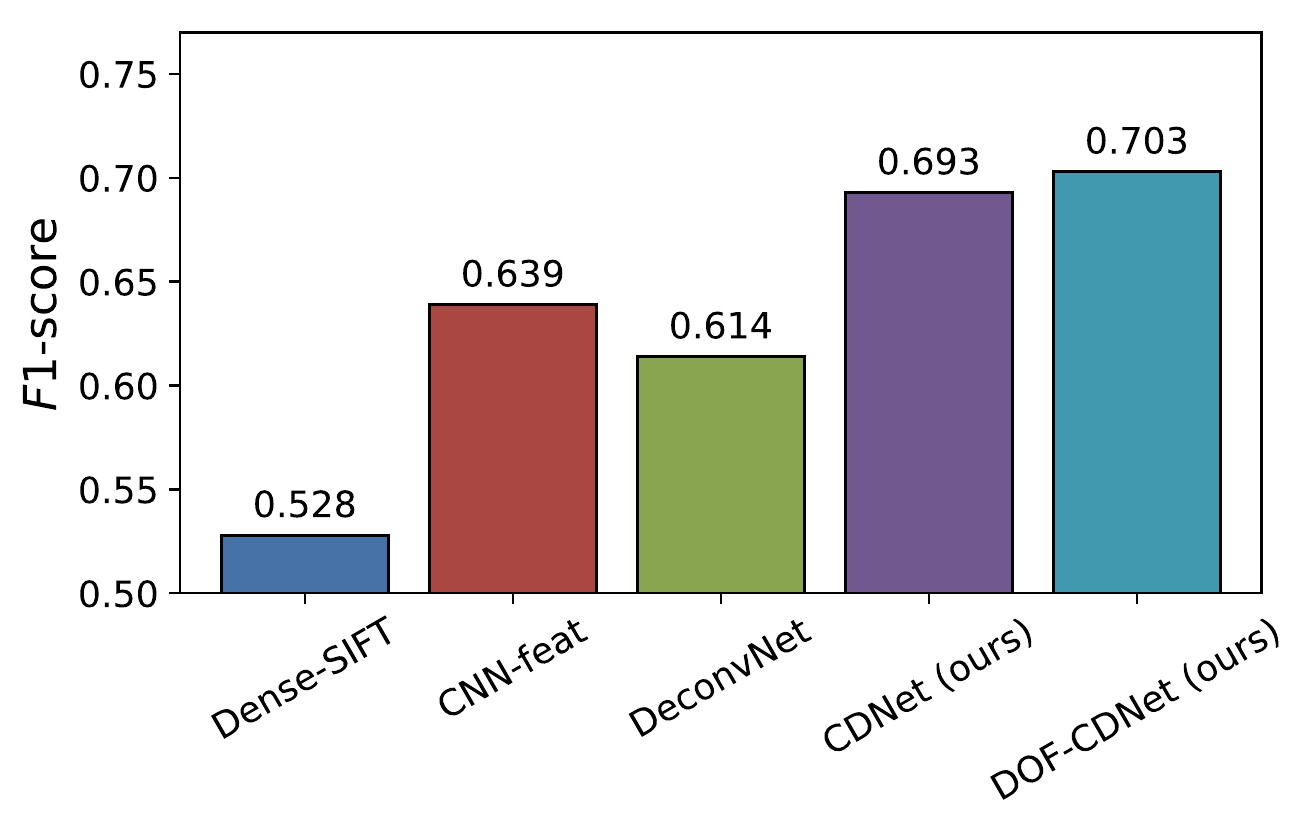}  
&\hspace*{0mm}\includegraphics[width=55mm,bb=0 30 372 241]{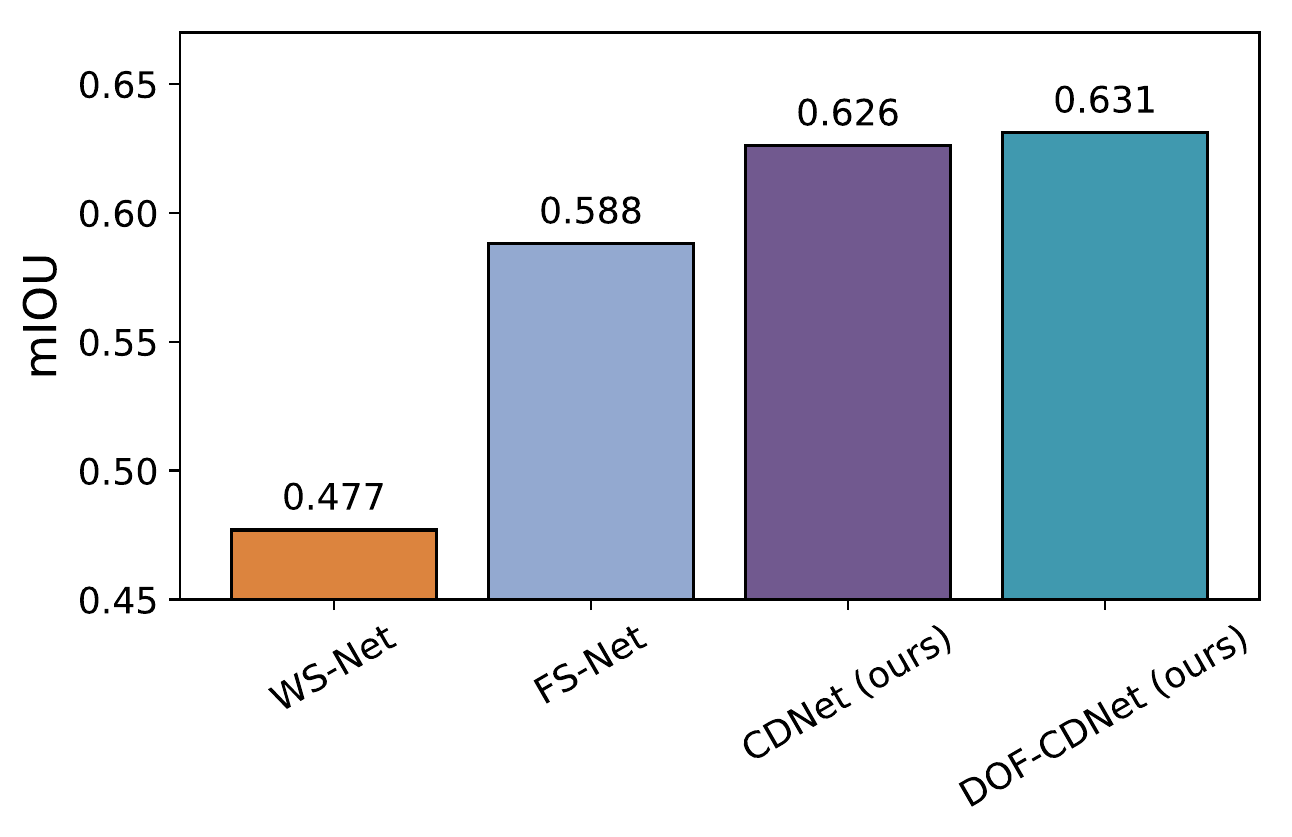} \\
\end{array}$
$\begin{array}{p{50mm}p{50mm}p{50mm}}
 \hspace*{11mm}\raisebox{1mm}{\scriptsize $F_{1}$ score of TSUNAMI}
&\hspace*{18mm}\raisebox{1mm}{\scriptsize $F_{1}$ score of GSV}
&\hspace*{15mm}\raisebox{1mm}{\scriptsize mIOU of TSUANAMI and GSV}
 \end{array}$

\caption{Estimation accuracy of change detection. The proposed methods, CDNet and DOF-CDNet, are compared with Dense-SIFT, CNN-feature \cite{sakurada2015change}, and DeconvNet \cite{alcantarilla2016street}, based methods using $F_1$ score, weakly supervised networks (WS-Net) and fully supervised networks (FS-Net) \cite{khan2017learning}, using mIOU.}
\label{fig:accuracy}

\end{center}

\end{figure*}

\subsection{Network Architecture}
The network architecture of DOF-CDNet is based on U-net \cite{ronneberger2015u,pix2pix}, which is one of state-of-the-art segmentation networks (Table \ref{tab:CNN_Structure}).
C, B, R, and D represent the layers of convolution, batch normalization, ReLU, and dropout, respectively.
From left to right, the numbers in parentheses indicate the number of layers, spatial filter size and stride amount of convolution filters, respectively.
All of the ReLUs in encoder are ``Leaky ReLU."

It is difficult to generate the ground-truth optical flow for real-scene imagery. 
Moreover, there are errors in the optical-flow estimated in Section \ref{subsec:optical_flow}. 
Therefore, in this study, the estimated optical flow vector $(f_u,f_v)$ is exploited as the input and its estimation error is modeled from the training data.

Images, $I$ and $I'$, captured at times, $t_0$ and $t_1$, and the optical flow image, $I_f$, are concatenated in the channel direction and are input as an eight-channel image. 
Each pixel value is normalized in $[-1,1]$.
The change mask as the ground-truth, $I_c$, is given to the output of the network as training data in the grayscale image ranging in $[0,s_\mathrm{max}]$ \footnote{We set $s_\mathrm{max}=255$ through all the experiments in this paper.}．
The $L1$ loss function $\mathcal{L}_{L1}$ is defined as follows:
\begin{equation}
\mathcal{L}_{L1}=\frac{1}{HW}\sum_{v=1}^{H}\sum_{u=1}^{W}|I_{c}^{(u,v)}-\phi(I,I',I_f)^{(u,v)}|_1,
\end{equation}
where $\phi$ is the pixel value of the change mask estimated by the trained networks.
In the prediction step, change probability, $p_c(u,v)$, of each pixel, $(u,v)$, is calculated as
\begin{equation}
p_c(u,v)=\frac{\phi(I,I',I_f)^{(u,v)}}{s_\mathrm{max}}.
\end{equation}

\begin{figure*}[!t]
\begin{center}
\vspace{0mm}
$\begin{array}{p{55mm}p{55mm}p{55mm}}
\hspace*{0mm}\includegraphics[width=55mm,bb=0 100 1024 224]{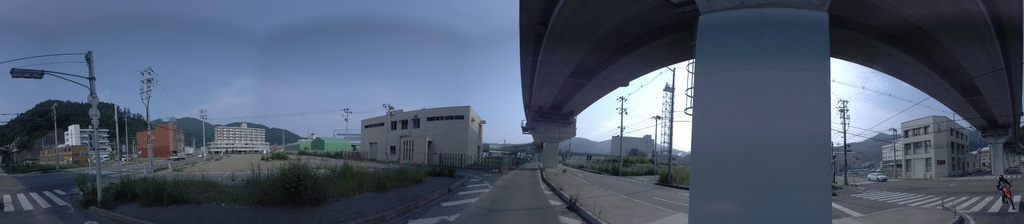} 
&\hspace*{0mm}\includegraphics[width=55mm,bb=0 100 1024 224]{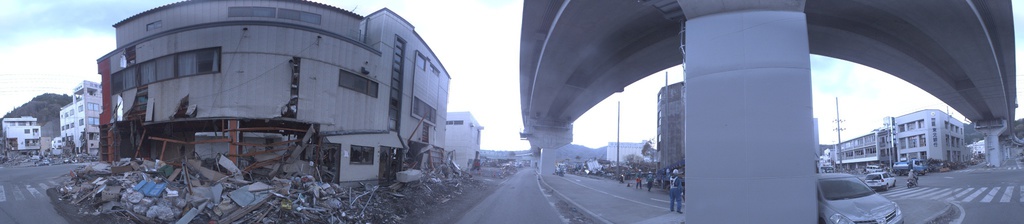}  
&\hspace*{0mm}\includegraphics[width=55mm,bb=0 100 1024 224]{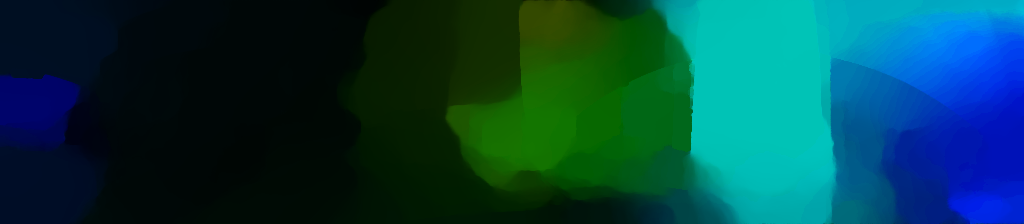} \\
\end{array}$
$\begin{array}{p{55mm}p{55mm}p{55mm}}
 \hspace*{18mm}\raisebox{1mm}{\scriptsize Input image of $t_0$}
&\hspace*{18mm}\raisebox{1mm}{\scriptsize Input image of $t_1$}
&\hspace*{16mm}\raisebox{1mm}{\scriptsize Estimated optical flow}
 \end{array}$
$\begin{array}{p{55mm}p{55mm}p{55mm}}
\hspace*{0mm}\includegraphics[width=55mm,bb=0 100 1024 224]{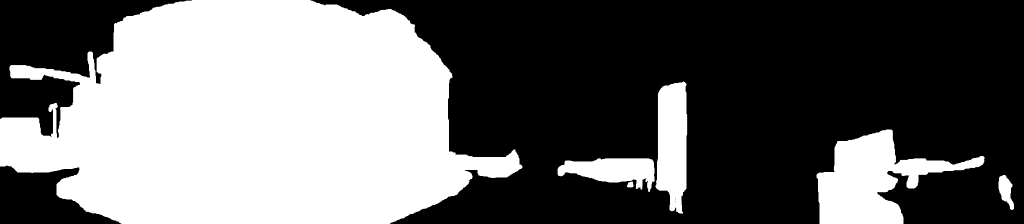} 
&\hspace*{0mm}\includegraphics[width=55mm,bb=0 100 1024 224]{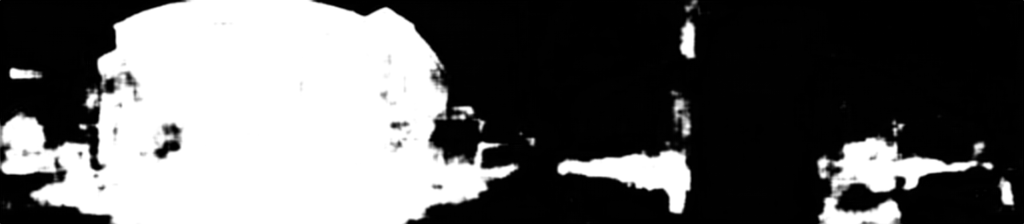} 
&\hspace*{0mm}\includegraphics[width=55mm,bb=0 100 1024 224]{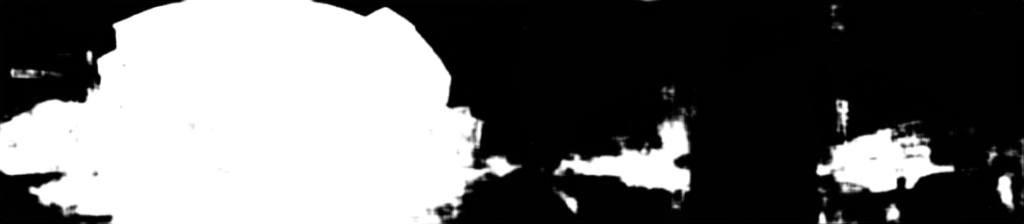}  \\
\end{array}$

$\begin{array}{p{55mm}p{55mm}p{55mm}}
\hspace*{14mm}\raisebox{1mm}{\scriptsize Hand-labeled ground-truth}
&\hspace*{13mm}\raisebox{1mm}{\scriptsize CDNet ($F_{1}$-score $= 0.9237$)}
&\hspace*{11mm}\raisebox{1mm}{\scriptsize DOF-CDNet ($F_{1}$-score $= 0.9314$)}
\end{array}$

\caption{Example of scene change detection of TSUNAMI.}
\label{fig:change_ex_1}

\end{center}
\end{figure*}

\begin{figure*}[!t]
\begin{center}
\vspace{0mm}
$\begin{array}{p{55mm}p{55mm}p{55mm}}
\hspace*{0mm}\includegraphics[width=55mm,bb=0 100 1024 224]{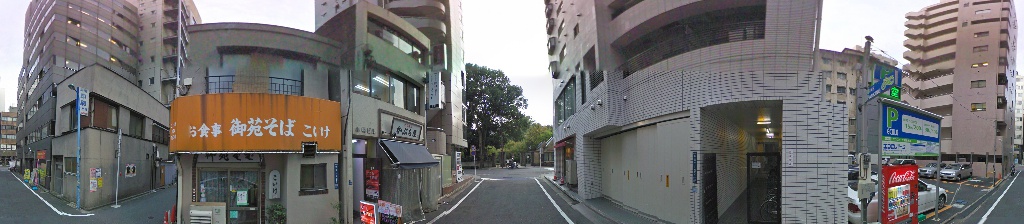} 
&\hspace*{0mm}\includegraphics[width=55mm,bb=0 100 1024 224]{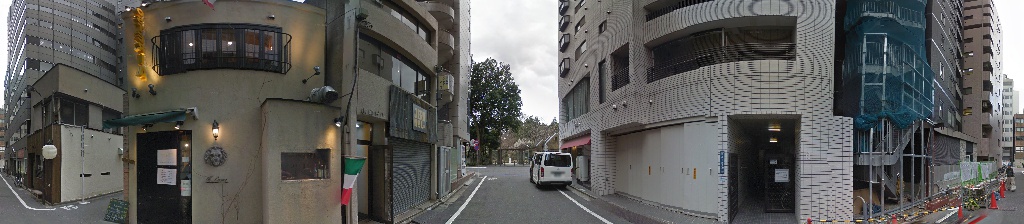}  
&\hspace*{0mm}\includegraphics[width=55mm,bb=0 100 1024 224]{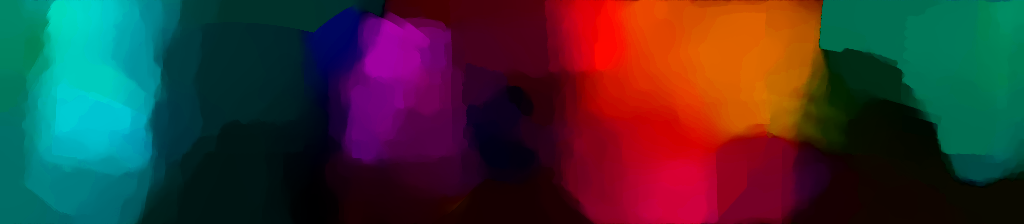} \\
\end{array}$
$\begin{array}{p{55mm}p{55mm}p{55mm}}
 \hspace*{18mm}\raisebox{1mm}{\scriptsize Input image of $t_0$}
&\hspace*{18mm}\raisebox{1mm}{\scriptsize Input image of $t_1$}
&\hspace*{16mm}\raisebox{1mm}{\scriptsize Estimated optical flow}
 \end{array}$
$\begin{array}{p{55mm}p{55mm}p{55mm}}
\hspace*{0mm}\includegraphics[width=55mm,bb=0 100 1024 224]{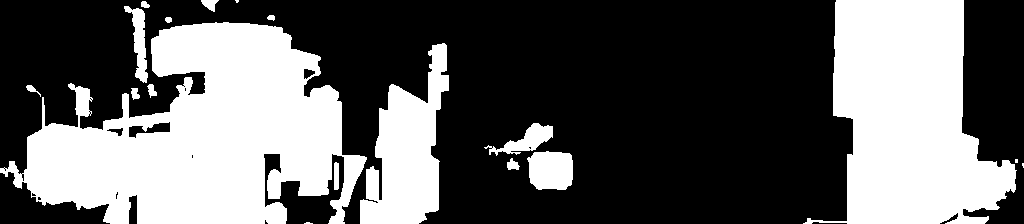} 
&\hspace*{0mm}\includegraphics[width=55mm,bb=0 100 1024 224]{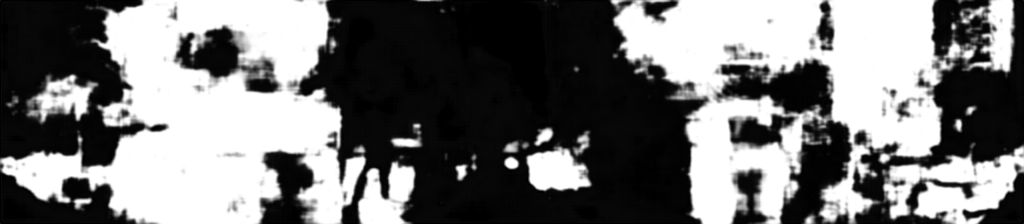} 
&\hspace*{0mm}\includegraphics[width=55mm,bb=0 100 1024 224]{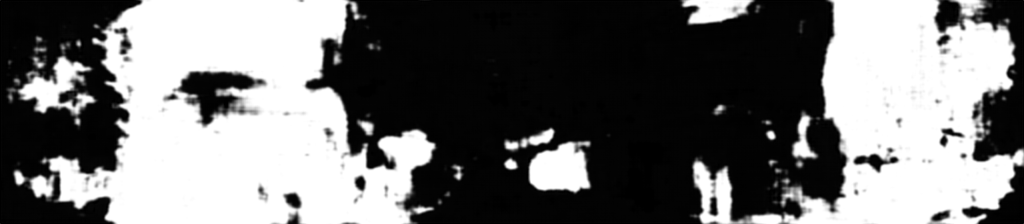}  \\
\end{array}$

$\begin{array}{p{55mm}p{55mm}p{55mm}}
\hspace*{14mm}\raisebox{1mm}{\scriptsize Hand-labeled ground-truth}
&\hspace*{13mm}\raisebox{1mm}{\scriptsize CDNet ($F_{1}$-score $= 0.6408$)}
&\hspace*{11mm}\raisebox{1mm}{\scriptsize DOF-CDNet ($F_{1}$-score $= 0.7721$)}
\end{array}$

\caption{Example of scene change detection of GSV.}
\label{fig:change_ex_2}

\end{center}

\end{figure*}

\begin{figure*}[!t]
\begin{center}

$\begin{array}{p{55mm}p{55mm}p{55mm}}
\hspace*{0mm}\includegraphics[width=55mm,bb=0 100 1024 224]{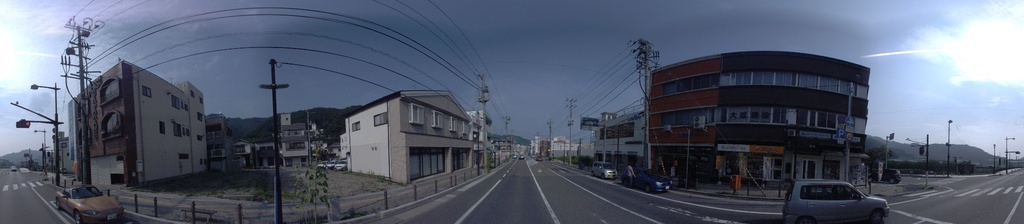} 
&\hspace*{0mm}\includegraphics[width=55mm,bb=0 100 1024 224]{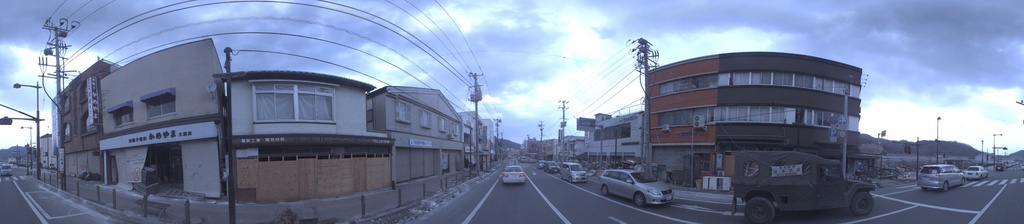}  
&\hspace*{0mm}\includegraphics[width=55mm,bb=0 100 1024 224]{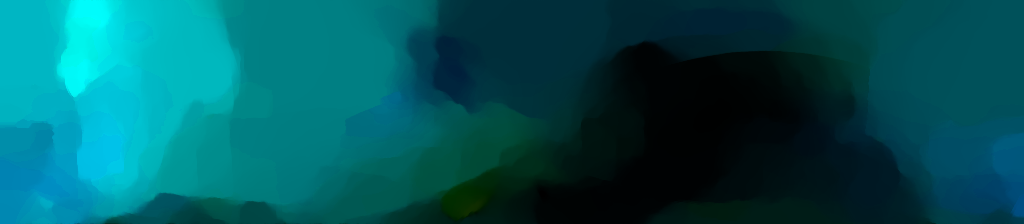} \\
\end{array}$
$\begin{array}{p{55mm}p{55mm}p{55mm}}
 \hspace*{18mm}\raisebox{1mm}{\scriptsize Input image of $t_0$}
&\hspace*{18mm}\raisebox{1mm}{\scriptsize Input image of $t_1$}
&\hspace*{16mm}\raisebox{1mm}{\scriptsize Estimated optical flow}
 \end{array}$
$\begin{array}{p{55mm}p{55mm}p{55mm}}
\hspace*{0mm}\includegraphics[width=55mm,bb=0 100 1024 224]{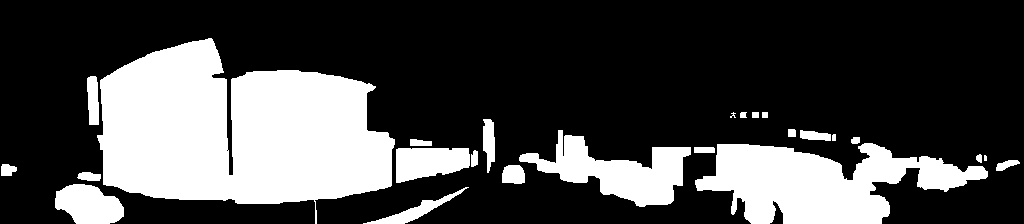} 
&\hspace*{0mm}\includegraphics[width=55mm,bb=0 100 1024 224]{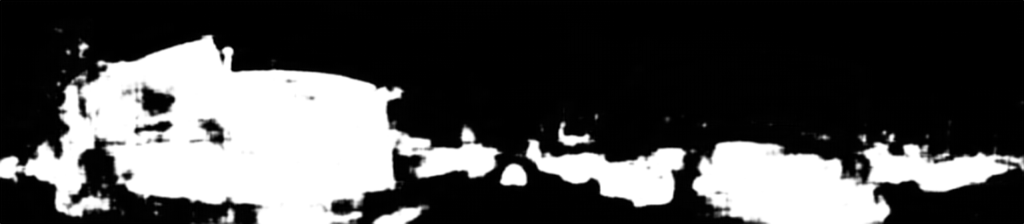} 
&\hspace*{0mm}\includegraphics[width=55mm,bb=0 100 1024 224]{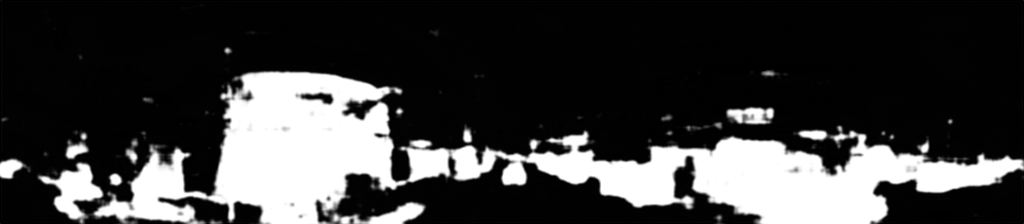}  \\
\end{array}$

$\begin{array}{p{55mm}p{55mm}p{55mm}}
\hspace*{14mm}\raisebox{1mm}{\scriptsize Hand-labeled ground-truth}
&\hspace*{13mm}\raisebox{1mm}{\scriptsize CDNet ($F_{1}$-score $= 0.8481$)}
&\hspace*{11mm}\raisebox{1mm}{\scriptsize DOF-CDNet ($F_{1}$-score $= 0.7217$)}
\end{array}$

\vspace{5mm}
$\begin{array}{p{55mm}p{55mm}p{55mm}}
\hspace*{0mm}\includegraphics[width=55mm,bb=0 100 1024 224]{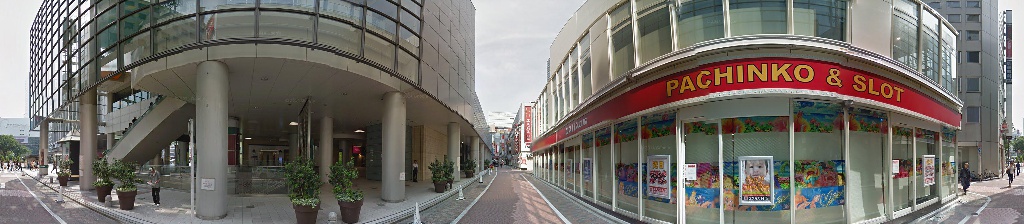} 
&\hspace*{0mm}\includegraphics[width=55mm,bb=0 100 1024 224]{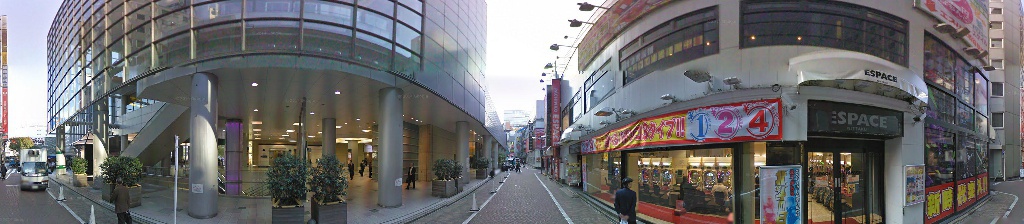}  
&\hspace*{0mm}\includegraphics[width=55mm,bb=0 100 1024 224]{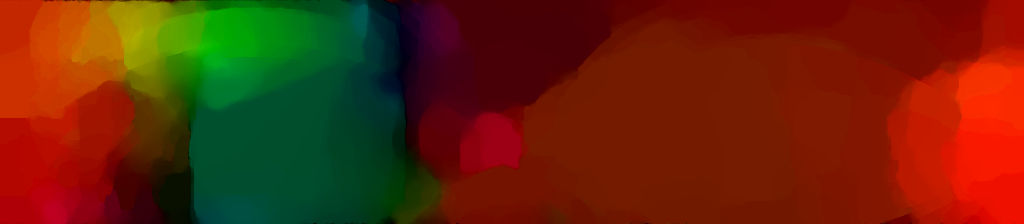} \\
\end{array}$
$\begin{array}{p{55mm}p{55mm}p{55mm}}
 \hspace*{18mm}\raisebox{1mm}{\scriptsize Input image of $t_0$}
&\hspace*{18mm}\raisebox{1mm}{\scriptsize Input image of $t_1$}
&\hspace*{16mm}\raisebox{1mm}{\scriptsize Estimated optical flow}
 \end{array}$
$\begin{array}{p{55mm}p{55mm}p{55mm}}
\hspace*{0mm}\includegraphics[width=55mm,bb=0 100 1024 224]{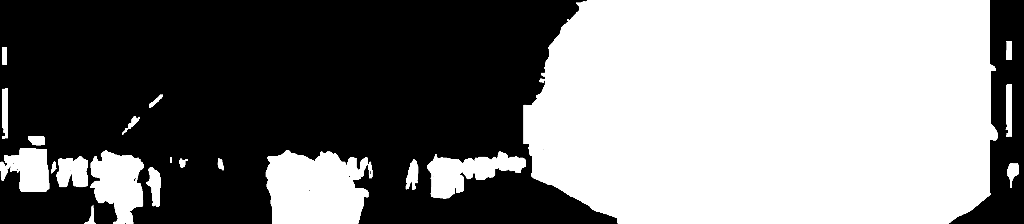} 
&\hspace*{0mm}\includegraphics[width=55mm,bb=0 100 1024 224]{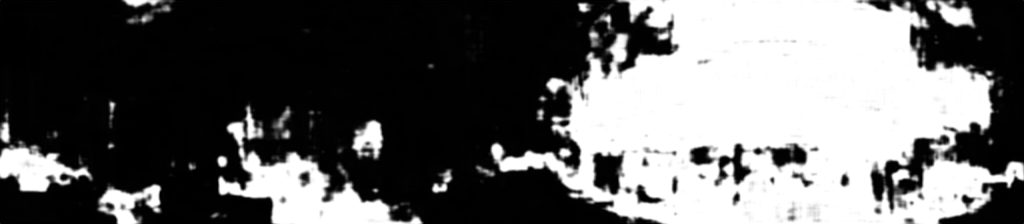} 
&\hspace*{0mm}\includegraphics[width=55mm,bb=0 100 1024 224]{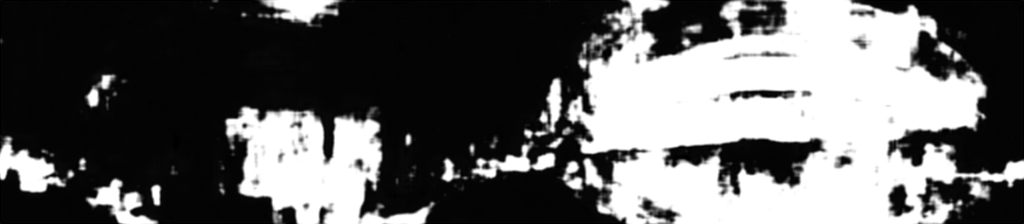}  \\
\end{array}$

$\begin{array}{p{55mm}p{55mm}p{55mm}}
\hspace*{14mm}\raisebox{1mm}{\scriptsize Hand-labeled ground-truth}
&\hspace*{13mm}\raisebox{1mm}{\scriptsize CDNet ($F_{1}$-score $= 0.8291$)}
&\hspace*{11mm}\raisebox{1mm}{\scriptsize DOF-CDNet ($F_{1}$-score $= 0.6644$)}
\end{array}$

\caption{Failure cases of the proposed method.}
\label{fig:change_ex_fail}

\end{center}
\end{figure*}

\section{Evaluation}
To evaluate the effectiveness of the proposed method, we conducted the experiments using the PCD dataset.
We compared the change detection network (CDNet), whose input is only a scene's image pair, and DOF-CDNet, whose input is both an image pair and the optical flow image.

The PCD dataset is composed of panoramic image pairs, $I$ and $I'$, each $1024\times224$, taken at two different time points, $t_0$ and $t_1$, and the change mask, $I_c$.
The subsets, TSUNAMI and GSV, each contains 100 image pairs.
First, the optical flow image, $I_f$ is estimated from $I$ and $I'$ by the method described in section \ref{subsec:optical_flow}.
Next, from the image set, $[I,I',I_f,I_c]$, patch images, each $224\times224$, are cropped sliding with 56 pixels width and resized to $256\times256$.
Furthermore, data augmentation was performed by rotating the patches.
Thus, 10,400 sets of image patches are generated.

\begin{figure*}[!t]
\begin{center}
\vspace{0mm}
$\begin{array}{p{55mm}p{55mm}p{55mm}}
\hspace*{0mm}\includegraphics[width=55mm,bb=0 100 1024 224]{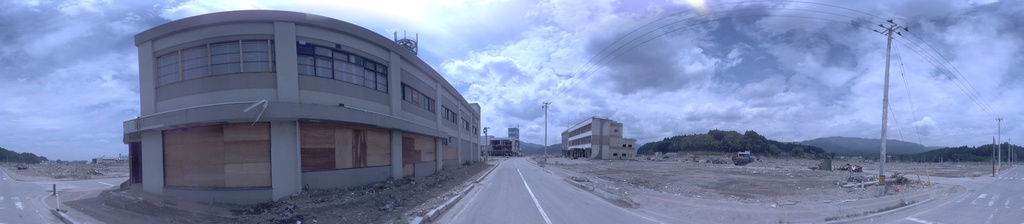} 
&\hspace*{0mm}\includegraphics[width=55mm,bb=0 100 1024 224]{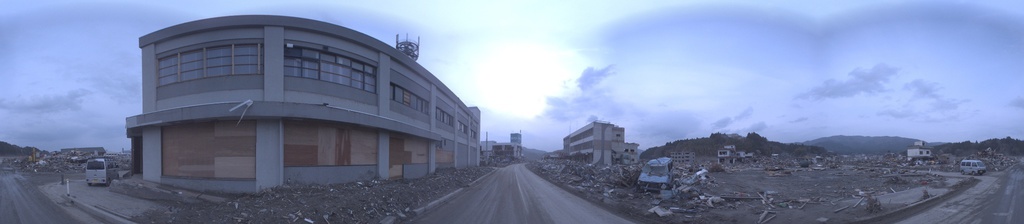}  
&\hspace*{0mm}\includegraphics[width=55mm,bb=0 100 1024 224]{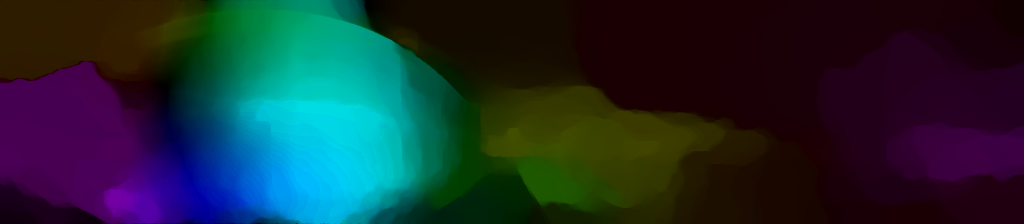} \\
\end{array}$
$\begin{array}{p{55mm}p{55mm}p{55mm}}
 \hspace*{18mm}\raisebox{1mm}{\scriptsize Input image of $t_0$}
&\hspace*{18mm}\raisebox{1mm}{\scriptsize Input image of $t_1$}
&\hspace*{16mm}\raisebox{1mm}{\scriptsize Estimated optical flow}
 \end{array}$
$\begin{array}{p{55mm}p{55mm}p{55mm}}
\hspace*{0mm}\includegraphics[width=55mm,bb=0 100 1024 224]{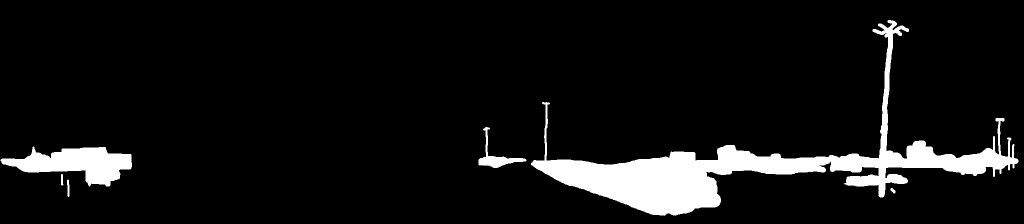} 
&\hspace*{0mm}\includegraphics[width=55mm,bb=0 100 1024 224]{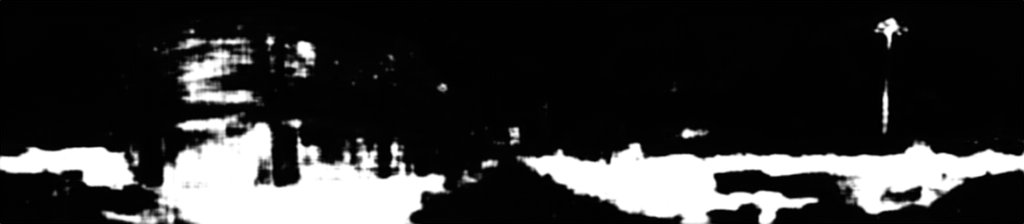} 
&\hspace*{0mm}\includegraphics[width=55mm,bb=0 100 1024 224]{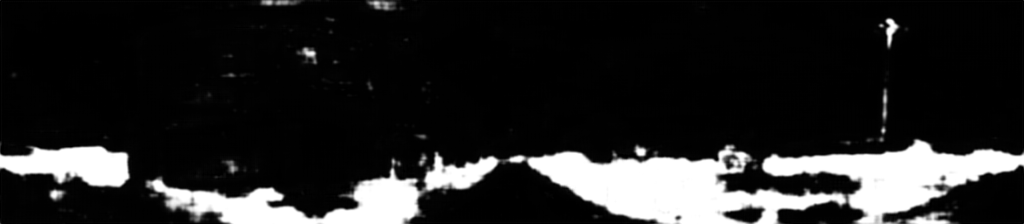}  \\
\end{array}$

$\begin{array}{p{55mm}p{55mm}p{55mm}}
\hspace*{14mm}\raisebox{1mm}{\scriptsize Hand-labeled ground-truth}
&\hspace*{13mm}\raisebox{1mm}{\scriptsize CDNet ($F_{1}$-score $= 0.5071$)}
&\hspace*{11mm}\raisebox{1mm}{\scriptsize DOF-CDNet ($F_{1}$-score $= 0.5312$)}
\end{array}$

\vspace{3mm}
$\begin{array}{p{55mm}p{55mm}p{55mm}}
\hspace*{0mm}\includegraphics[width=55mm,bb=0 100 1024 224]{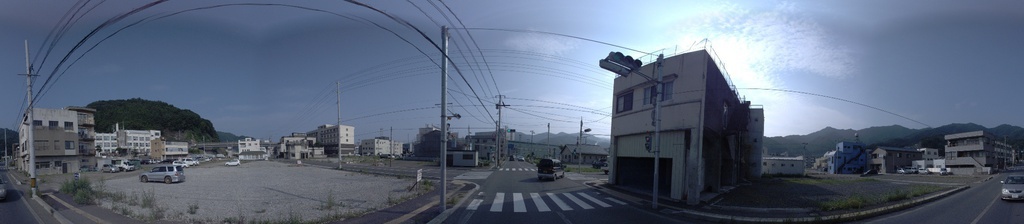} 
&\hspace*{0mm}\includegraphics[width=55mm,bb=0 100 1024 224]{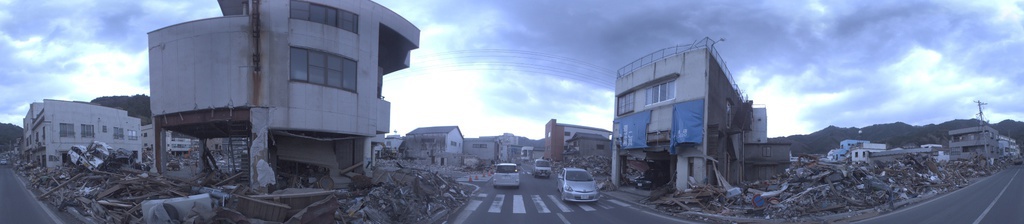}  
&\hspace*{0mm}\includegraphics[width=55mm,bb=0 100 1024 224]{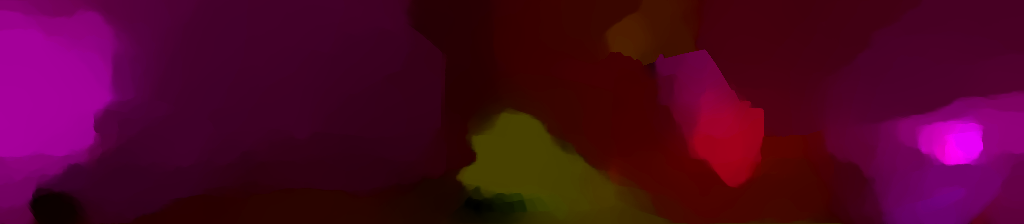} \\
\end{array}$
$\begin{array}{p{55mm}p{55mm}p{55mm}}
 \hspace*{18mm}\raisebox{1mm}{\scriptsize Input image of $t_0$}
&\hspace*{18mm}\raisebox{1mm}{\scriptsize Input image of $t_1$}
&\hspace*{16mm}\raisebox{1mm}{\scriptsize Estimated optical flow}
 \end{array}$
$\begin{array}{p{55mm}p{55mm}p{55mm}}
\hspace*{0mm}\includegraphics[width=55mm,bb=0 100 1024 224]{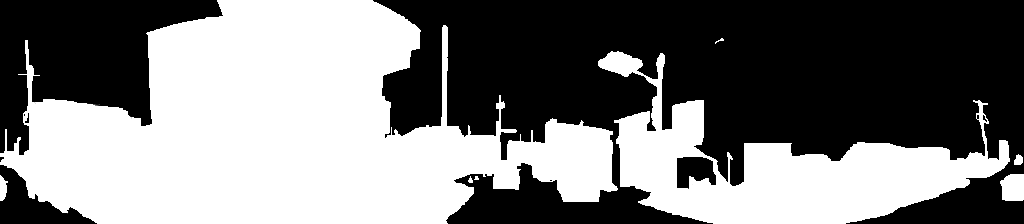} 
&\hspace*{0mm}\includegraphics[width=55mm,bb=0 100 1024 224]{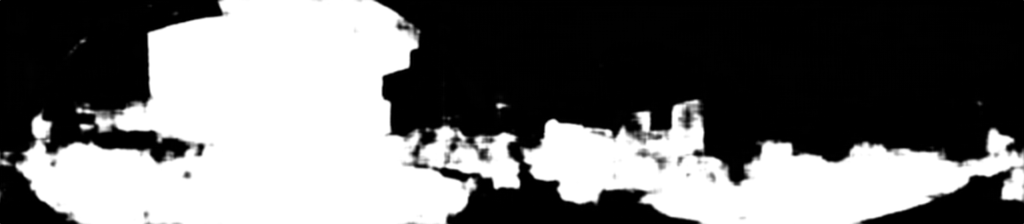} 
&\hspace*{0mm}\includegraphics[width=55mm,bb=0 100 1024 224]{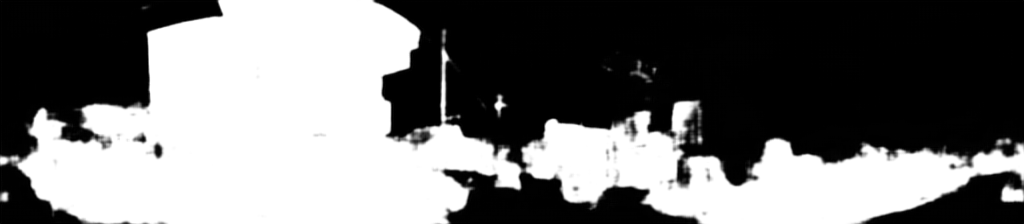}  \\
\end{array}$

$\begin{array}{p{55mm}p{55mm}p{55mm}}
\hspace*{14mm}\raisebox{1mm}{\scriptsize Hand-labeled ground-truth}
&\hspace*{13mm}\raisebox{1mm}{\scriptsize CDNet ($F_{1}$-score $= 0.9237$)}
&\hspace*{11mm}\raisebox{1mm}{\scriptsize DOF-CDNet ($F_{1}$-score $= 0.9314$)}
\end{array}$

\vspace{3mm}
$\begin{array}{p{55mm}p{55mm}p{55mm}}
\hspace*{0mm}\includegraphics[width=55mm,bb=0 100 1024 224]{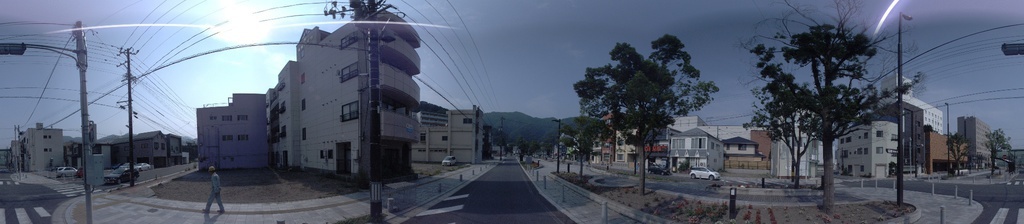} 
&\hspace*{0mm}\includegraphics[width=55mm,bb=0 100 1024 224]{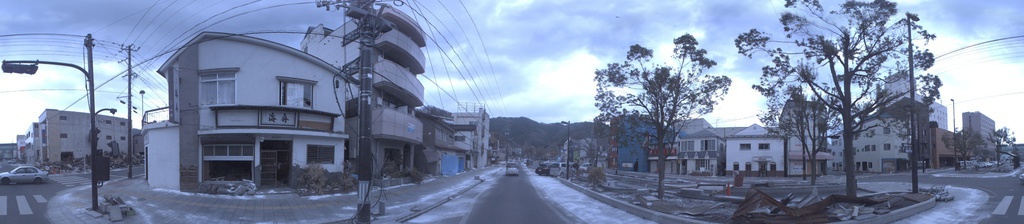}  
&\hspace*{0mm}\includegraphics[width=55mm,bb=0 100 1024 224]{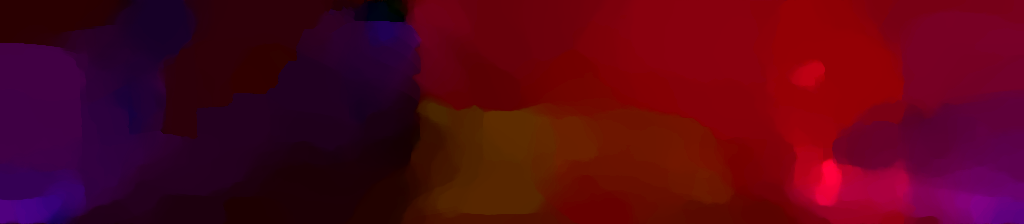} \\
\end{array}$
$\begin{array}{p{55mm}p{55mm}p{55mm}}
 \hspace*{18mm}\raisebox{1mm}{\scriptsize Input image of $t_0$}
&\hspace*{18mm}\raisebox{1mm}{\scriptsize Input image of $t_1$}
&\hspace*{16mm}\raisebox{1mm}{\scriptsize Estimated optical flow}
 \end{array}$
$\begin{array}{p{55mm}p{55mm}p{55mm}}
\hspace*{0mm}\includegraphics[width=55mm,bb=0 100 1024 224]{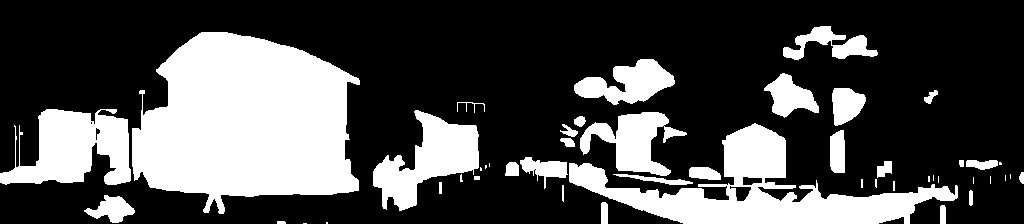} 
&\hspace*{0mm}\includegraphics[width=55mm,bb=0 100 1024 224]{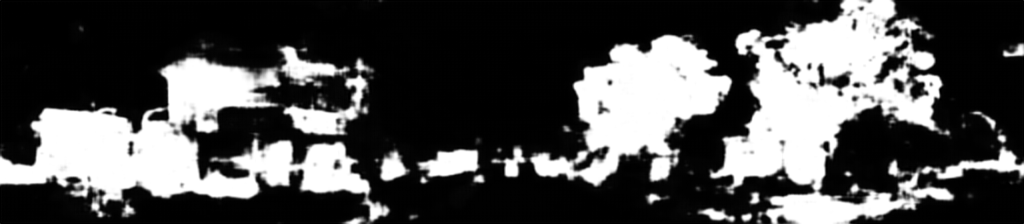} 
&\hspace*{0mm}\includegraphics[width=55mm,bb=0 100 1024 224]{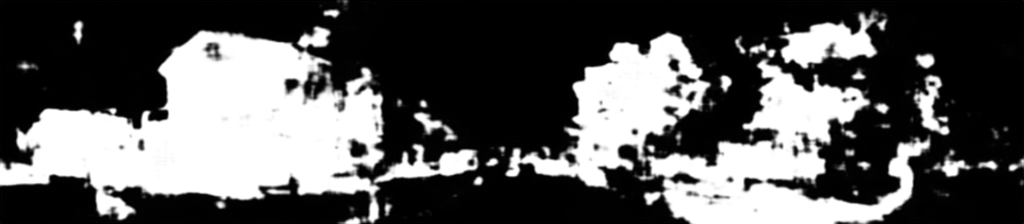}  \\
\end{array}$

$\begin{array}{p{55mm}p{55mm}p{55mm}}
\hspace*{14mm}\raisebox{1mm}{\scriptsize Hand-labeled ground-truth}
&\hspace*{13mm}\raisebox{1mm}{\scriptsize CDNet ($F_{1}$-score $= 0.5995$)}
&\hspace*{11mm}\raisebox{1mm}{\scriptsize DOF-CDNet ($F_{1}$-score $= 0.6954$)}
\end{array}$

\vspace{3mm}
$\begin{array}{p{55mm}p{55mm}p{55mm}}
\hspace*{0mm}\includegraphics[width=55mm,bb=0 100 1024 224]{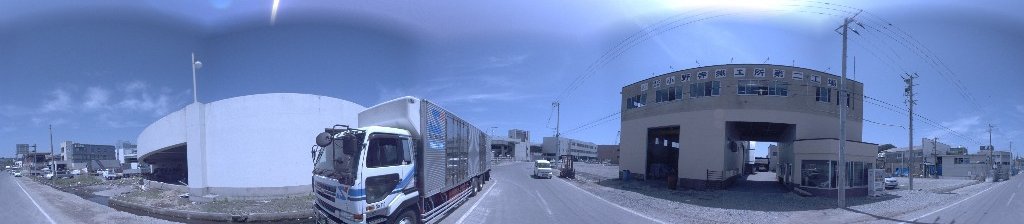} 
&\hspace*{0mm}\includegraphics[width=55mm,bb=0 100 1024 224]{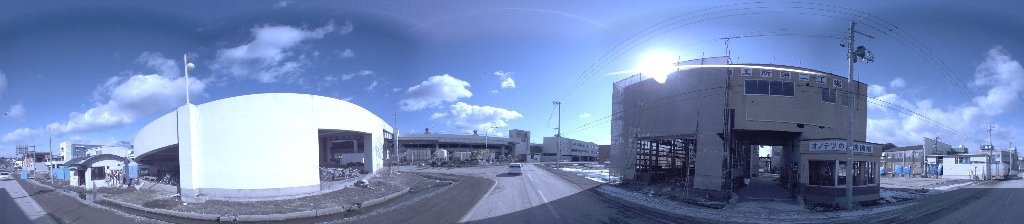}  
&\hspace*{0mm}\includegraphics[width=55mm,bb=0 100 1024 224]{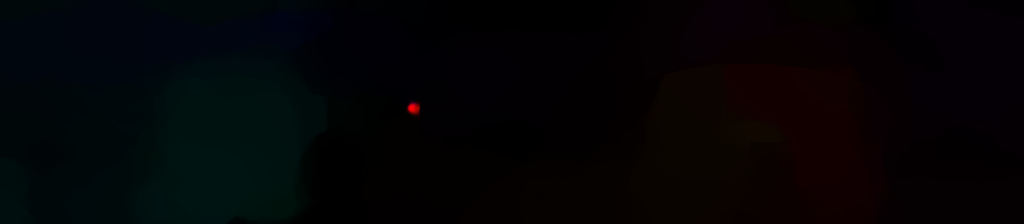} \\
\end{array}$
$\begin{array}{p{55mm}p{55mm}p{55mm}}
 \hspace*{18mm}\raisebox{1mm}{\scriptsize Input image of $t_0$}
&\hspace*{18mm}\raisebox{1mm}{\scriptsize Input image of $t_1$}
&\hspace*{16mm}\raisebox{1mm}{\scriptsize Estimated optical flow}
 \end{array}$
$\begin{array}{p{55mm}p{55mm}p{55mm}}
\hspace*{0mm}\includegraphics[width=55mm,bb=0 100 1024 224]{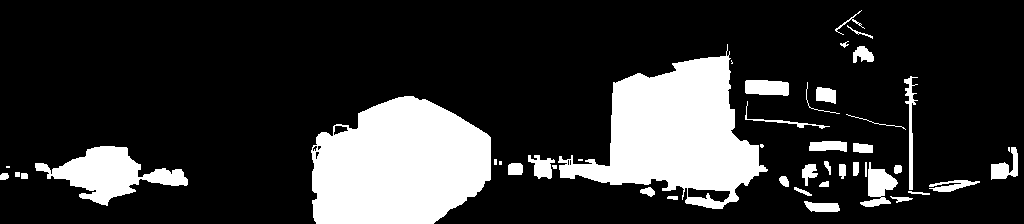} 
&\hspace*{0mm}\includegraphics[width=55mm,bb=0 100 1024 224]{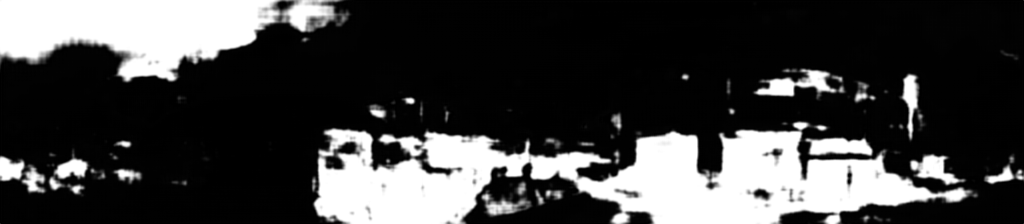} 
&\hspace*{0mm}\includegraphics[width=55mm,bb=0 100 1024 224]{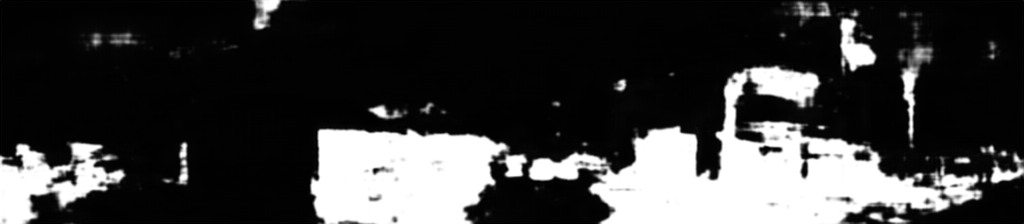}  \\
\end{array}$

$\begin{array}{p{55mm}p{55mm}p{55mm}}
\hspace*{14mm}\raisebox{1mm}{\scriptsize Hand-labeled ground-truth}
&\hspace*{13mm}\raisebox{1mm}{\scriptsize CDNet ($F_{1}$-score $= 0.4575$)}
&\hspace*{11mm}\raisebox{1mm}{\scriptsize DOF-CDNet ($F_{1}$-score $= 0.5437$)}
\end{array}$

\vspace{3mm}
$\begin{array}{p{55mm}p{55mm}p{55mm}}
\hspace*{0mm}\includegraphics[width=55mm,bb=0 100 1024 224]{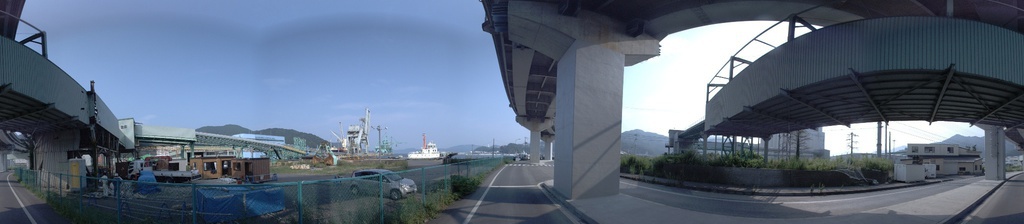} 
&\hspace*{0mm}\includegraphics[width=55mm,bb=0 100 1024 224]{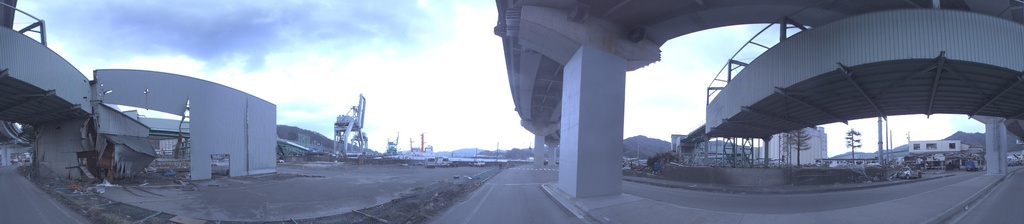}  
&\hspace*{0mm}\includegraphics[width=55mm,bb=0 100 1024 224]{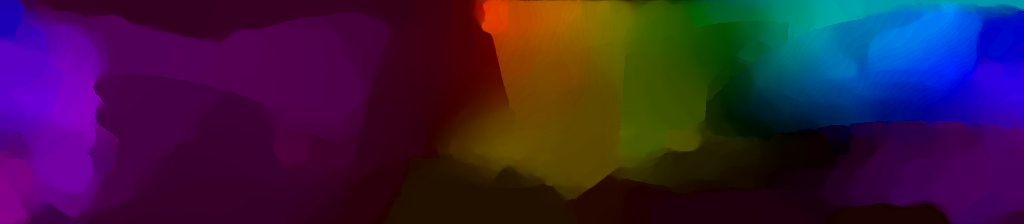} \\
\end{array}$
$\begin{array}{p{55mm}p{55mm}p{55mm}}
 \hspace*{18mm}\raisebox{1mm}{\scriptsize Input image of $t_0$}
&\hspace*{18mm}\raisebox{1mm}{\scriptsize Input image of $t_1$}
&\hspace*{16mm}\raisebox{1mm}{\scriptsize Estimated optical flow}
 \end{array}$
$\begin{array}{p{55mm}p{55mm}p{55mm}}
\hspace*{0mm}\includegraphics[width=55mm,bb=0 100 1024 224]{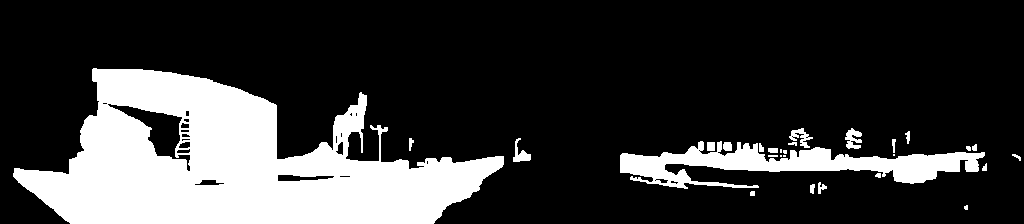} 
&\hspace*{0mm}\includegraphics[width=55mm,bb=0 100 1024 224]{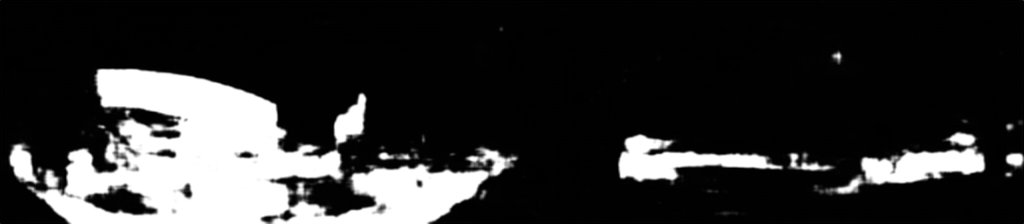} 
&\hspace*{0mm}\includegraphics[width=55mm,bb=0 100 1024 224]{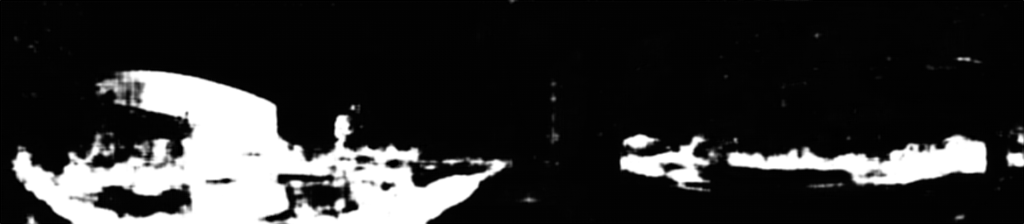}  \\
\end{array}$

$\begin{array}{p{55mm}p{55mm}p{55mm}}
\hspace*{14mm}\raisebox{1mm}{\scriptsize Hand-labeled ground-truth}
&\hspace*{13mm}\raisebox{1mm}{\scriptsize CDNet ($F_{1}$-score $= 0.8160$)}
&\hspace*{11mm}\raisebox{1mm}{\scriptsize DOF-CDNet ($F_{1}$-score $= 0.8240$)}
\end{array}$

\vspace{3mm}
$\begin{array}{p{55mm}p{55mm}p{55mm}}
\hspace*{0mm}\includegraphics[width=55mm,bb=0 100 1024 224]{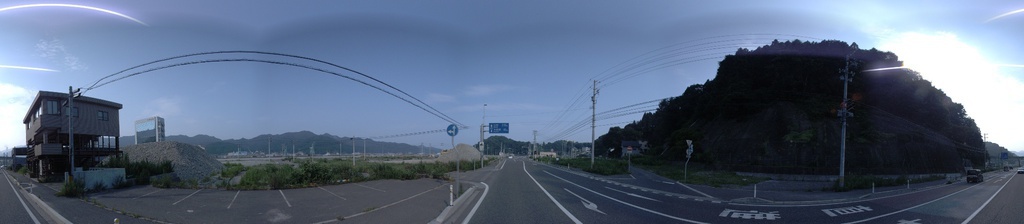} 
&\hspace*{0mm}\includegraphics[width=55mm,bb=0 100 1024 224]{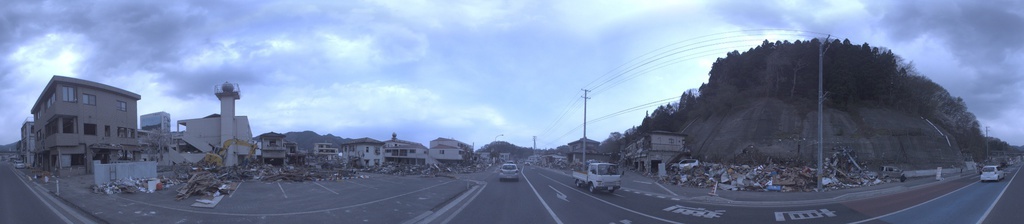}  
&\hspace*{0mm}\includegraphics[width=55mm,bb=0 100 1024 224]{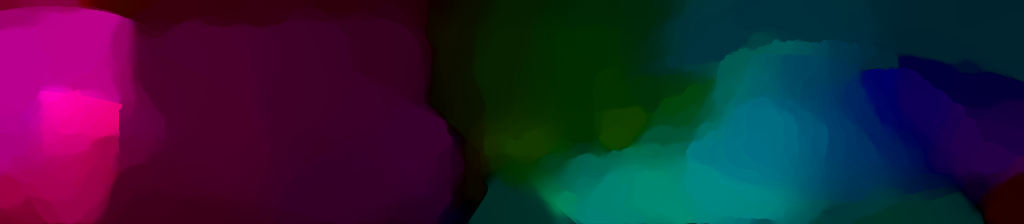} \\
\end{array}$
$\begin{array}{p{55mm}p{55mm}p{55mm}}
 \hspace*{18mm}\raisebox{1mm}{\scriptsize Input image of $t_0$}
&\hspace*{18mm}\raisebox{1mm}{\scriptsize Input image of $t_1$}
&\hspace*{16mm}\raisebox{1mm}{\scriptsize Estimated optical flow}
 \end{array}$
$\begin{array}{p{55mm}p{55mm}p{55mm}}
\hspace*{0mm}\includegraphics[width=55mm,bb=0 100 1024 224]{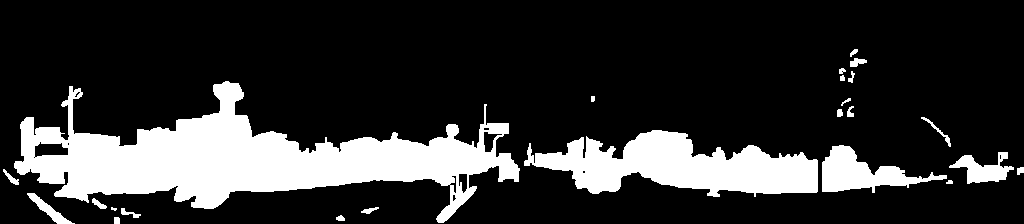} 
&\hspace*{0mm}\includegraphics[width=55mm,bb=0 100 1024 224]{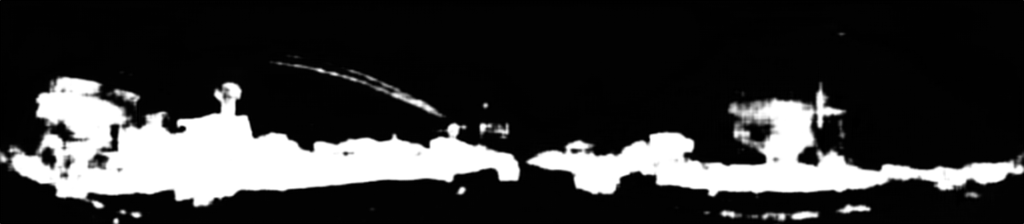} 
&\hspace*{0mm}\includegraphics[width=55mm,bb=0 100 1024 224]{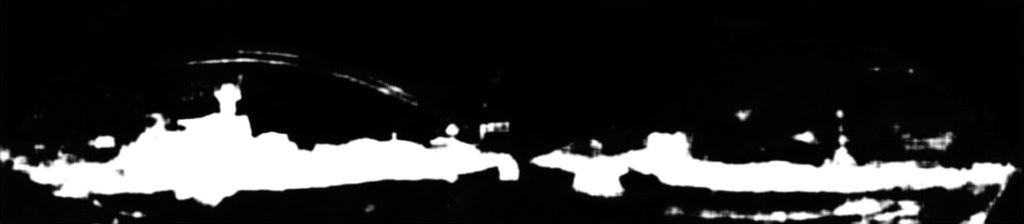}  \\
\end{array}$

$\begin{array}{p{55mm}p{55mm}p{55mm}}
\hspace*{14mm}\raisebox{1mm}{\scriptsize Hand-labeled ground-truth}
&\hspace*{13mm}\raisebox{1mm}{\scriptsize CDNet ($F_{1}$-score $= 0.8015$)}
&\hspace*{11mm}\raisebox{1mm}{\scriptsize DOF-CDNet ($F_{1}$-score $= 0.8102$)}
\end{array}$

 \caption{Additional results of scene change detection of TSUNAMI.}
\label{fig:change_ex_TSUNAMI}

\end{center}
\end{figure*}

\begin{figure*}[!t]
\begin{center}
\vspace{0mm}
$\begin{array}{p{55mm}p{55mm}p{55mm}}
\hspace*{0mm}\includegraphics[width=55mm,bb=0 100 1024 224]{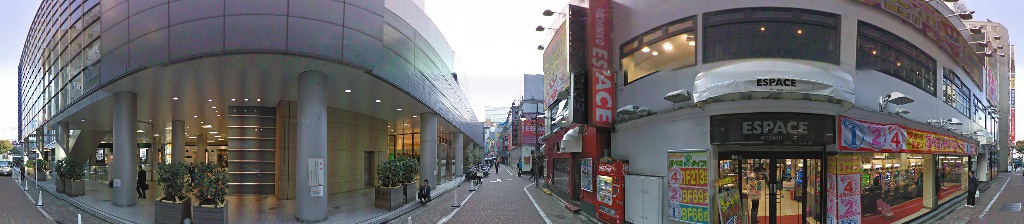} 
&\hspace*{0mm}\includegraphics[width=55mm,bb=0 100 1024 224]{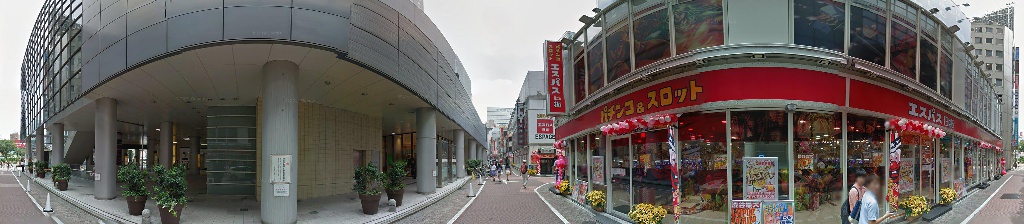}  
&\hspace*{0mm}\includegraphics[width=55mm,bb=0 100 1024 224]{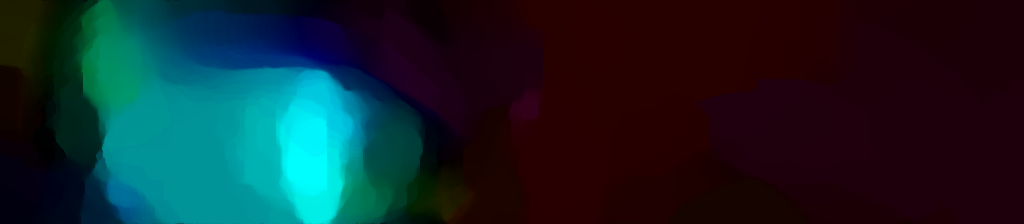} \\
\end{array}$
$\begin{array}{p{55mm}p{55mm}p{55mm}}
 \hspace*{18mm}\raisebox{1mm}{\scriptsize Input image of $t_0$}
&\hspace*{18mm}\raisebox{1mm}{\scriptsize Input image of $t_1$}
&\hspace*{16mm}\raisebox{1mm}{\scriptsize Estimated optical flow}
 \end{array}$
$\begin{array}{p{55mm}p{55mm}p{55mm}}
\hspace*{0mm}\includegraphics[width=55mm,bb=0 100 1024 224]{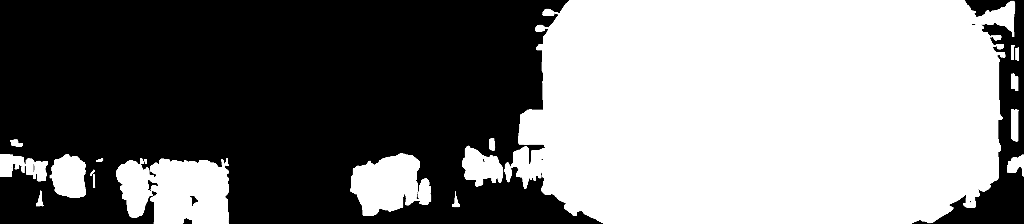} 
&\hspace*{0mm}\includegraphics[width=55mm,bb=0 100 1024 224]{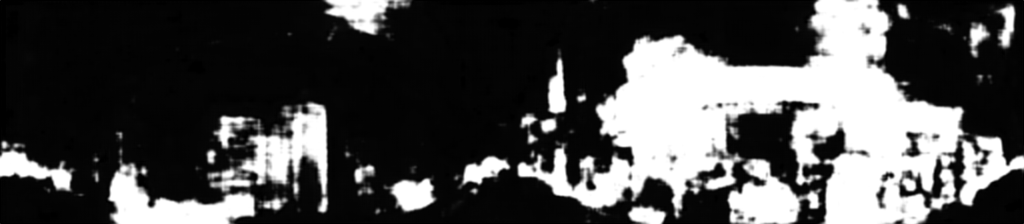} 
&\hspace*{0mm}\includegraphics[width=55mm,bb=0 100 1024 224]{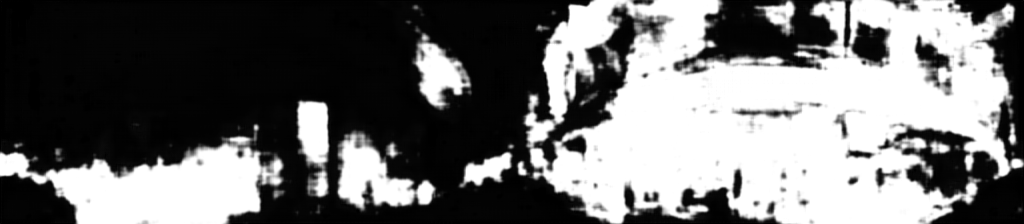}  \\
\end{array}$

$\begin{array}{p{55mm}p{55mm}p{55mm}}
\hspace*{14mm}\raisebox{1mm}{\scriptsize Hand-labeled ground-truth}
&\hspace*{13mm}\raisebox{1mm}{\scriptsize CDNet ($F_{1}$-score $= 0.6326$)}
&\hspace*{11mm}\raisebox{1mm}{\scriptsize DOF-CDNet ($F_{1}$-score $= 0.7395$)}
\end{array}$

\vspace{3mm}
$\begin{array}{p{55mm}p{55mm}p{55mm}}
\hspace*{0mm}\includegraphics[width=55mm,bb=0 100 1024 224]{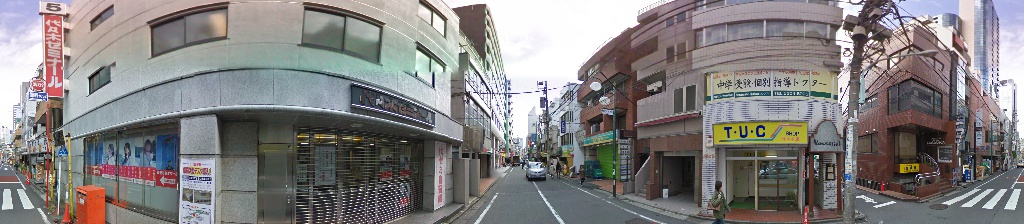} 
&\hspace*{0mm}\includegraphics[width=55mm,bb=0 100 1024 224]{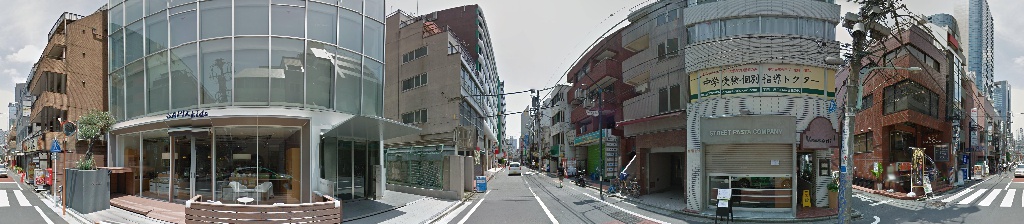}  
&\hspace*{0mm}\includegraphics[width=55mm,bb=0 100 1024 224]{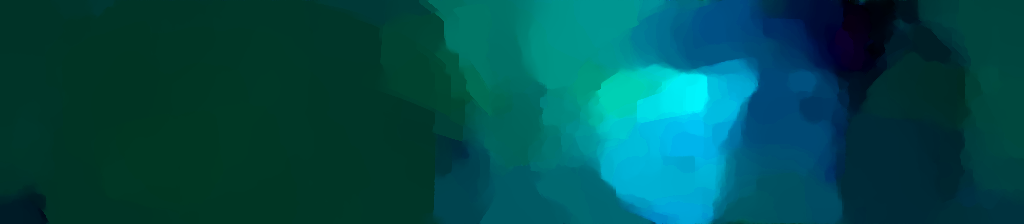} \\
\end{array}$
$\begin{array}{p{55mm}p{55mm}p{55mm}}
 \hspace*{18mm}\raisebox{1mm}{\scriptsize Input image of $t_0$}
&\hspace*{18mm}\raisebox{1mm}{\scriptsize Input image of $t_1$}
&\hspace*{16mm}\raisebox{1mm}{\scriptsize Estimated optical flow}
 \end{array}$
$\begin{array}{p{55mm}p{55mm}p{55mm}}
\hspace*{0mm}\includegraphics[width=55mm,bb=0 100 1024 224]{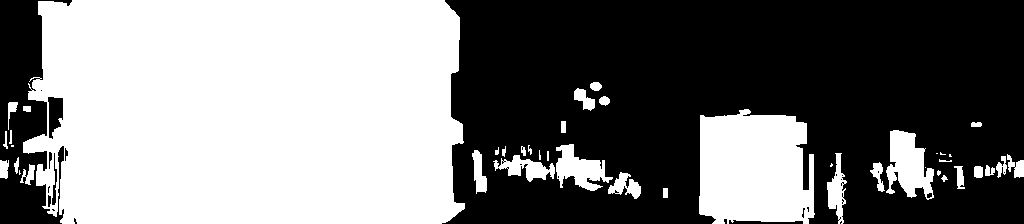} 
&\hspace*{0mm}\includegraphics[width=55mm,bb=0 100 1024 224]{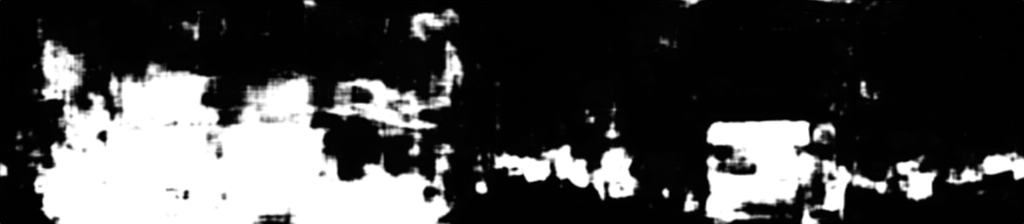} 
&\hspace*{0mm}\includegraphics[width=55mm,bb=0 100 1024 224]{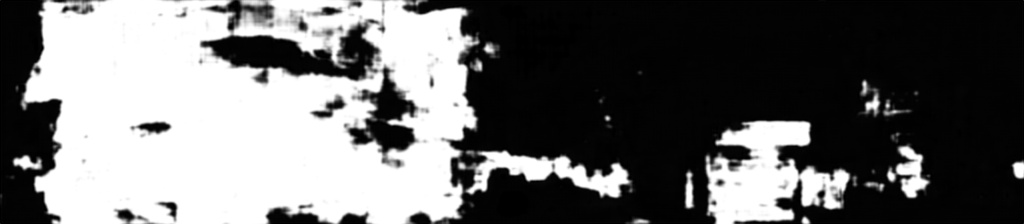}  \\
\end{array}$

$\begin{array}{p{55mm}p{55mm}p{55mm}}
\hspace*{14mm}\raisebox{1mm}{\scriptsize Hand-labeled ground-truth}
&\hspace*{13mm}\raisebox{1mm}{\scriptsize CDNet ($F_{1}$-score $= 0.6698$)}
&\hspace*{11mm}\raisebox{1mm}{\scriptsize DOF-CDNet ($F_{1}$-score $= 0.7944$)}
\end{array}$

\vspace{3mm}
$\begin{array}{p{55mm}p{55mm}p{55mm}}
\hspace*{0mm}\includegraphics[width=55mm,bb=0 100 1024 224]{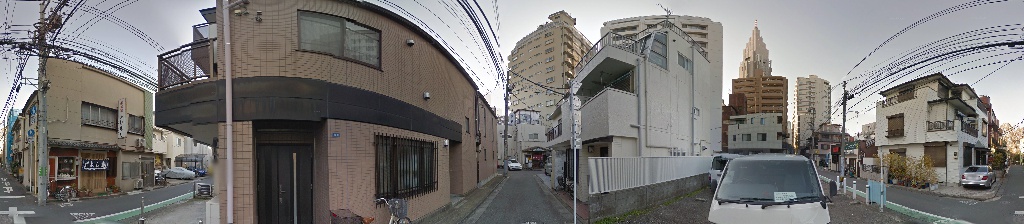} 
&\hspace*{0mm}\includegraphics[width=55mm,bb=0 100 1024 224]{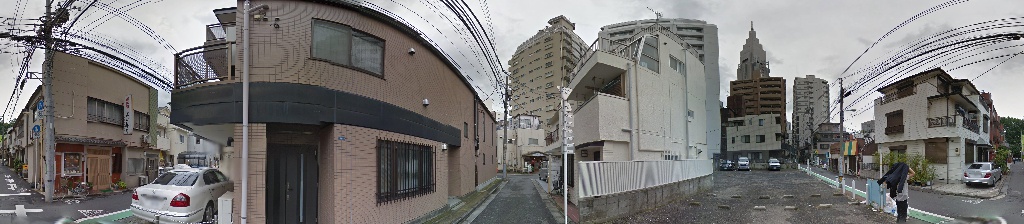}  
&\hspace*{0mm}\includegraphics[width=55mm,bb=0 100 1024 224]{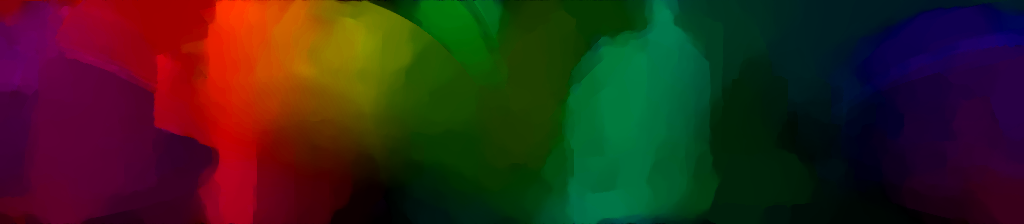} \\
\end{array}$
$\begin{array}{p{55mm}p{55mm}p{55mm}}
 \hspace*{18mm}\raisebox{1mm}{\scriptsize Input image of $t_0$}
&\hspace*{18mm}\raisebox{1mm}{\scriptsize Input image of $t_1$}
&\hspace*{16mm}\raisebox{1mm}{\scriptsize Estimated optical flow}
 \end{array}$
$\begin{array}{p{55mm}p{55mm}p{55mm}}
\hspace*{0mm}\includegraphics[width=55mm,bb=0 100 1024 224]{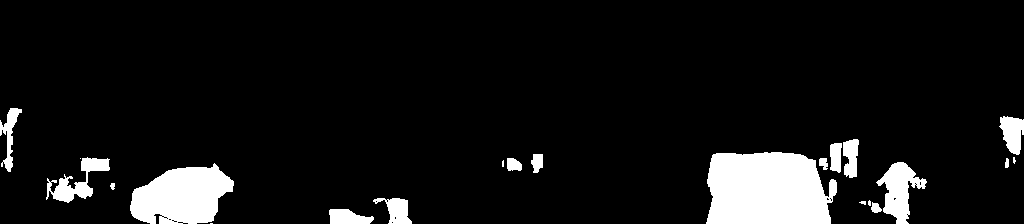} 
&\hspace*{0mm}\includegraphics[width=55mm,bb=0 100 1024 224]{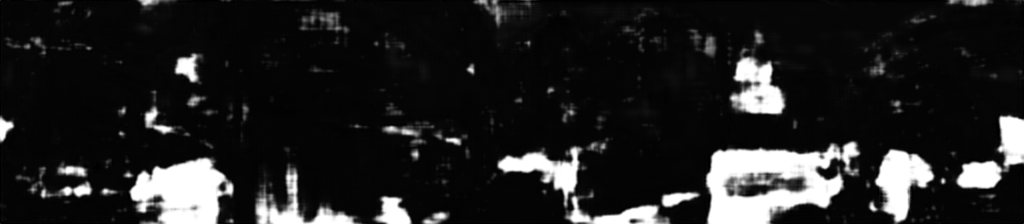} 
&\hspace*{0mm}\includegraphics[width=55mm,bb=0 100 1024 224]{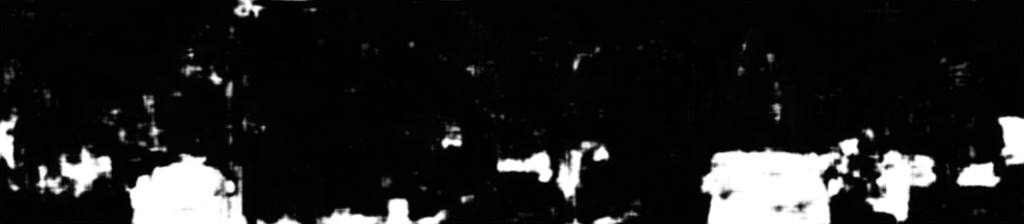}  \\
\end{array}$

$\begin{array}{p{55mm}p{55mm}p{55mm}}
\hspace*{14mm}\raisebox{1mm}{\scriptsize Hand-labeled ground-truth}
&\hspace*{13mm}\raisebox{1mm}{\scriptsize CDNet ($F_{1}$-score $= 0.6215$)}
&\hspace*{11mm}\raisebox{1mm}{\scriptsize DOF-CDNet ($F_{1}$-score $= 0.6415$)}
\end{array}$

\vspace{3mm}
$\begin{array}{p{55mm}p{55mm}p{55mm}}
\hspace*{0mm}\includegraphics[width=55mm,bb=0 100 1024 224]{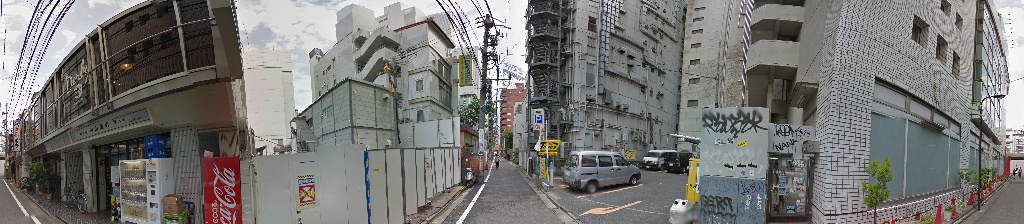} 
&\hspace*{0mm}\includegraphics[width=55mm,bb=0 100 1024 224]{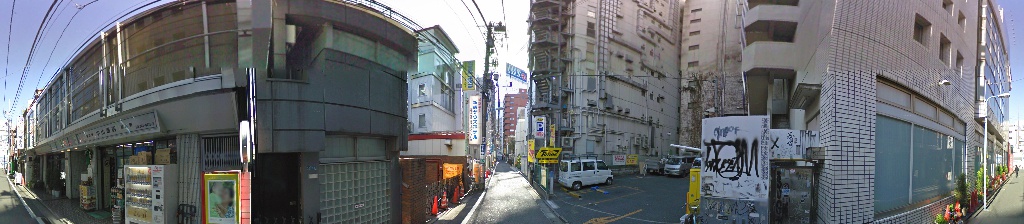}  
&\hspace*{0mm}\includegraphics[width=55mm,bb=0 100 1024 224]{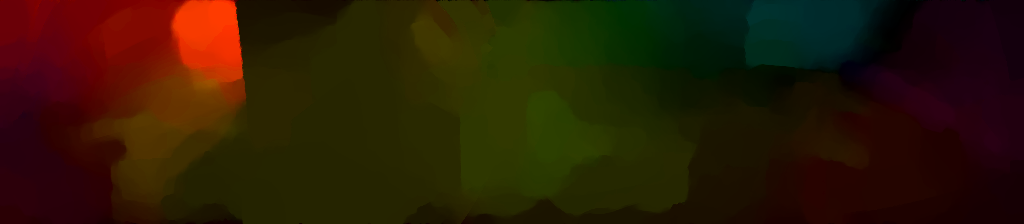} \\
\end{array}$
$\begin{array}{p{55mm}p{55mm}p{55mm}}
 \hspace*{18mm}\raisebox{1mm}{\scriptsize Input image of $t_0$}
&\hspace*{18mm}\raisebox{1mm}{\scriptsize Input image of $t_1$}
&\hspace*{16mm}\raisebox{1mm}{\scriptsize Estimated optical flow}
 \end{array}$
$\begin{array}{p{55mm}p{55mm}p{55mm}}
\hspace*{0mm}\includegraphics[width=55mm,bb=0 100 1024 224]{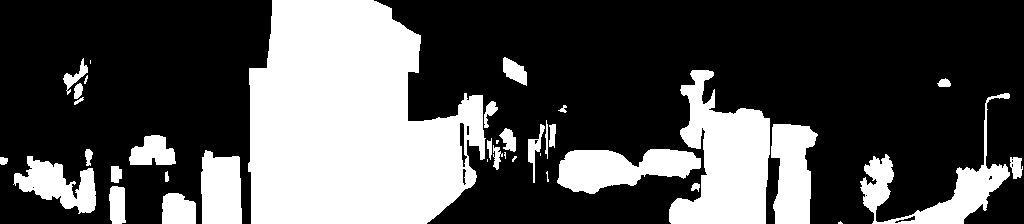} 
&\hspace*{0mm}\includegraphics[width=55mm,bb=0 100 1024 224]{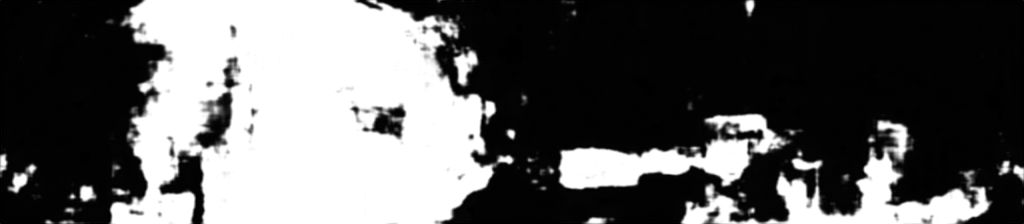} 
&\hspace*{0mm}\includegraphics[width=55mm,bb=0 100 1024 224]{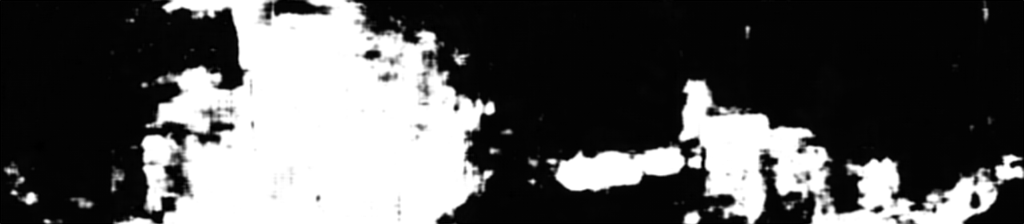}  \\
\end{array}$

$\begin{array}{p{55mm}p{55mm}p{55mm}}
\hspace*{14mm}\raisebox{1mm}{\scriptsize Hand-labeled ground-truth}
&\hspace*{13mm}\raisebox{1mm}{\scriptsize CDNet ($F_{1}$-score $= 0.7236$)}
&\hspace*{11mm}\raisebox{1mm}{\scriptsize DOF-CDNet ($F_{1}$-score $= 0.7401$)}
\end{array}$

\vspace{3mm}
$\begin{array}{p{55mm}p{55mm}p{55mm}}
\hspace*{0mm}\includegraphics[width=55mm,bb=0 100 1024 224]{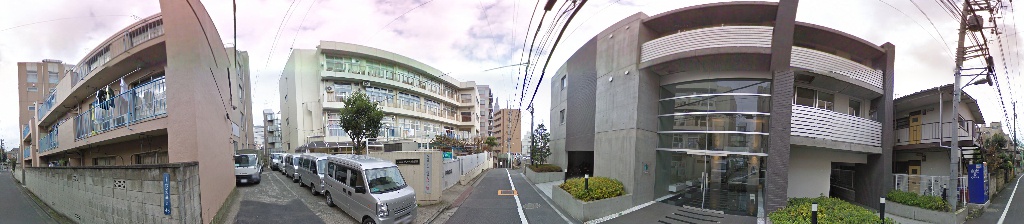} 
&\hspace*{0mm}\includegraphics[width=55mm,bb=0 100 1024 224]{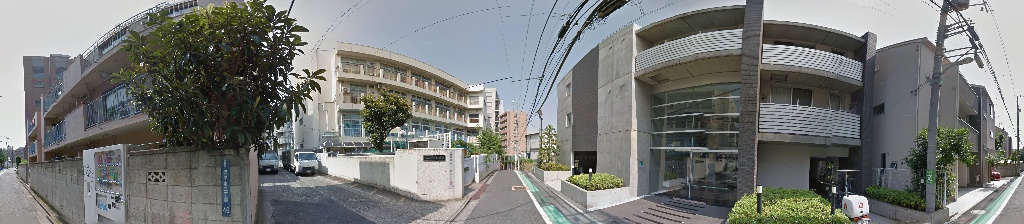}  
&\hspace*{0mm}\includegraphics[width=55mm,bb=0 100 1024 224]{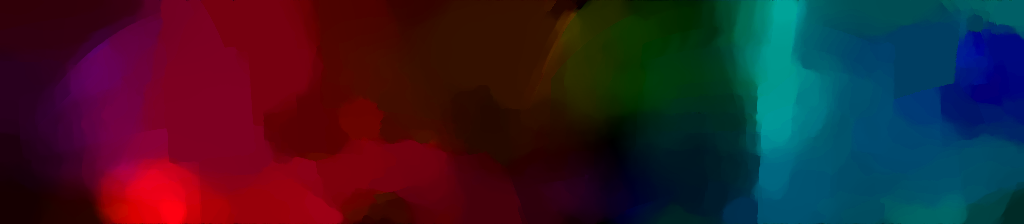} \\
\end{array}$
$\begin{array}{p{55mm}p{55mm}p{55mm}}
 \hspace*{18mm}\raisebox{1mm}{\scriptsize Input image of $t_0$}
&\hspace*{18mm}\raisebox{1mm}{\scriptsize Input image of $t_1$}
&\hspace*{16mm}\raisebox{1mm}{\scriptsize Estimated optical flow}
 \end{array}$
$\begin{array}{p{55mm}p{55mm}p{55mm}}
\hspace*{0mm}\includegraphics[width=55mm,bb=0 100 1024 224]{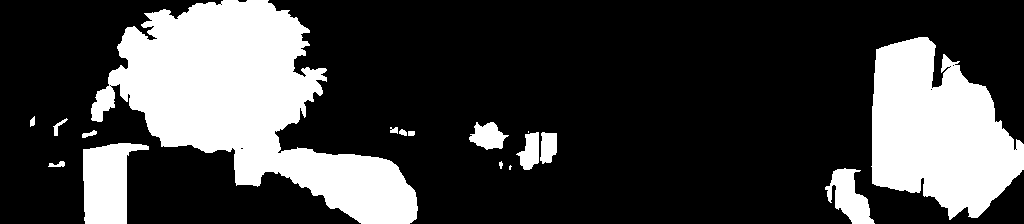} 
&\hspace*{0mm}\includegraphics[width=55mm,bb=0 100 1024 224]{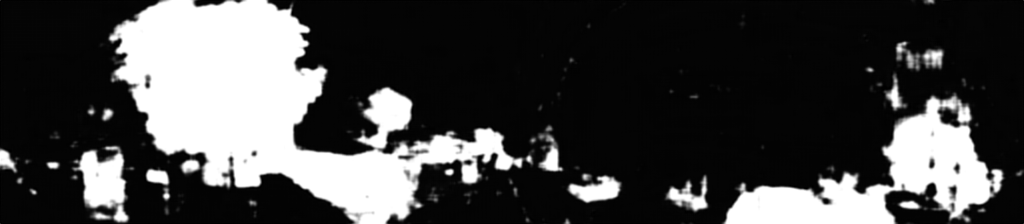} 
&\hspace*{0mm}\includegraphics[width=55mm,bb=0 100 1024 224]{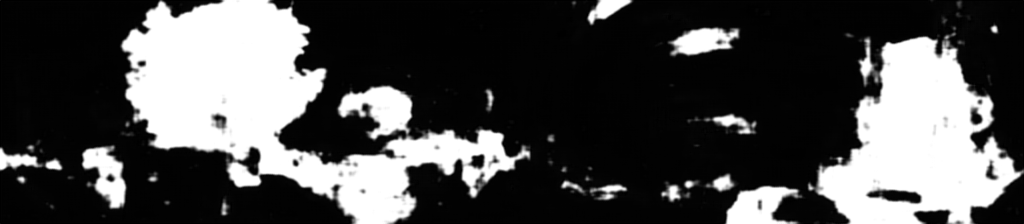}  \\
\end{array}$

$\begin{array}{p{55mm}p{55mm}p{55mm}}
\hspace*{14mm}\raisebox{1mm}{\scriptsize Hand-labeled ground-truth}
&\hspace*{13mm}\raisebox{1mm}{\scriptsize CDNet ($F_{1}$-score $= 0.7418$)}
&\hspace*{11mm}\raisebox{1mm}{\scriptsize DOF-CDNet ($F_{1}$-score $= 0.7660$)}
\end{array}$

\vspace{3mm}
$\begin{array}{p{55mm}p{55mm}p{55mm}}
\hspace*{0mm}\includegraphics[width=55mm,bb=0 100 1024 224]{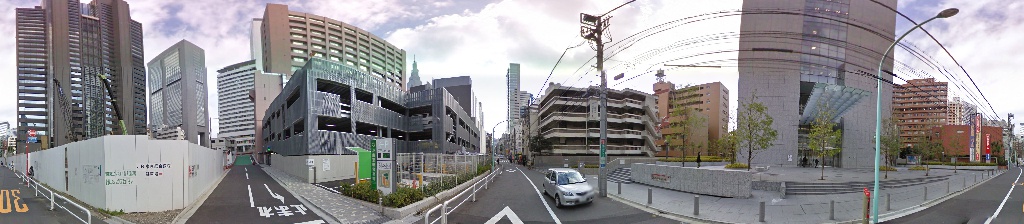} 
&\hspace*{0mm}\includegraphics[width=55mm,bb=0 100 1024 224]{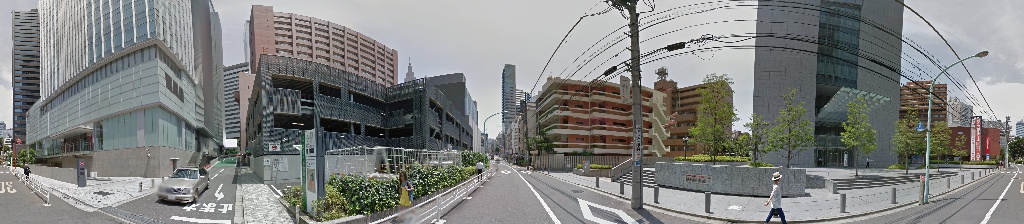}  
&\hspace*{0mm}\includegraphics[width=55mm,bb=0 100 1024 224]{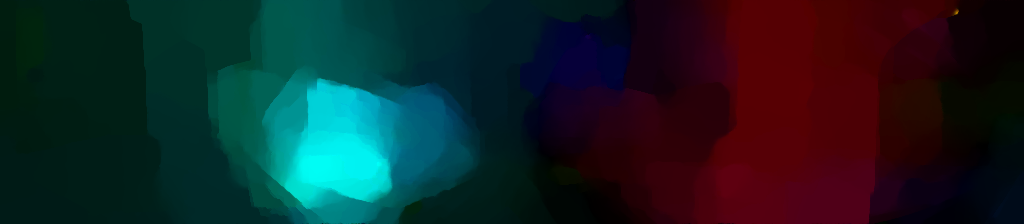} \\
\end{array}$
$\begin{array}{p{55mm}p{55mm}p{55mm}}
 \hspace*{18mm}\raisebox{1mm}{\scriptsize Input image of $t_0$}
&\hspace*{18mm}\raisebox{1mm}{\scriptsize Input image of $t_1$}
&\hspace*{16mm}\raisebox{1mm}{\scriptsize Estimated optical flow}
 \end{array}$
$\begin{array}{p{55mm}p{55mm}p{55mm}}
\hspace*{0mm}\includegraphics[width=55mm,bb=0 100 1024 224]{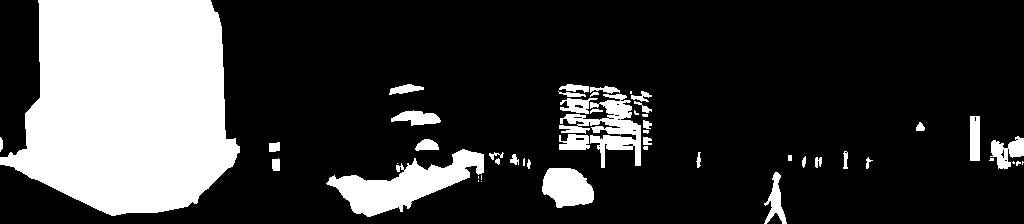} 
&\hspace*{0mm}\includegraphics[width=55mm,bb=0 100 1024 224]{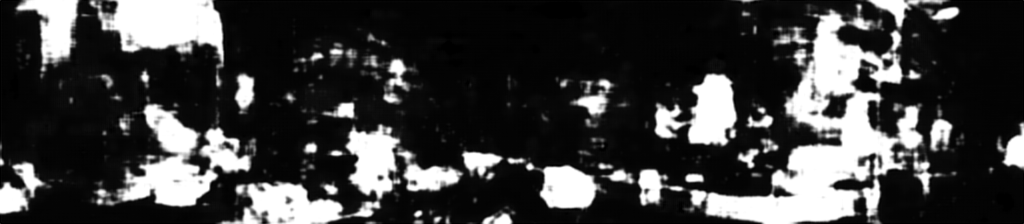} 
&\hspace*{0mm}\includegraphics[width=55mm,bb=0 100 1024 224]{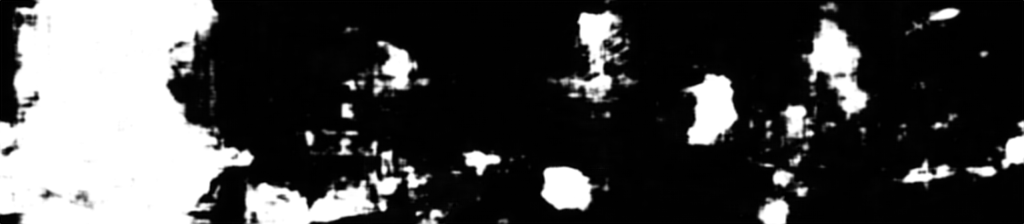}  \\
\end{array}$

$\begin{array}{p{55mm}p{55mm}p{55mm}}
\hspace*{14mm}\raisebox{1mm}{\scriptsize Hand-labeled ground-truth}
&\hspace*{13mm}\raisebox{1mm}{\scriptsize CDNet ($F_{1}$-score $= 0.3558$)}
&\hspace*{11mm}\raisebox{1mm}{\scriptsize DOF-CDNet ($F_{1}$-score $= 0.6483$)}
\end{array}$

\caption{Additional results of scene change detection of GSV.}
\label{fig:change_ex_GSV}

\end{center}

\end{figure*}

Estimation accuracies of the proposed methods are evaluated using the dataset through five-fold cross-validation \footnote{For five-hold cross-validation, the image patches are generated after each 100 image pairs of TSUNAMI and GSV are divided by the same ratio (Training : Test = 4 : 1), respectively.}.
Figure \ref{fig:accuracy} shows $F_{1}$ scores and mean intersection-over-union (mIOU) of each method.
Both CDNet and DOF-CDNet outperform the CNN grid feature-based method \cite{sakurada2015change}, the DeconvNet-based method \cite{alcantarilla2016street}, the weakly supervised networks, and the fully supervised networks \cite{khan2017learning}.
Furthermore, in the GSV dataset, whose changes are comparatively small, an optical flow based method (i.e., DOF-CDNet) is effective for reducing errors in large optical flow areas (see Figure \ref{fig:change_ex_1} and Figure \ref{fig:change_ex_2}). 
Figure \ref{fig:change_ex_fail} shows failure cases of the proposed method.
The change detection errors can be caused by the estimation error of the optical flow, especially for change area, and the lack of the training samples of the camera viewpoint changes.
Although CDNet outperforms DOF-CDNet in the cases, the proposed methods consistently outperform existing methods.
Figure \ref{fig:change_ex_TSUNAMI} and Figure \ref{fig:change_ex_GSV} show the additional results of the proposed method.

\section{Conclusion}
This paper proposed a CNN based on dense optical flow to introduce the robustness against the difference of camera viewpoints change between input images.
To estimate the optical flow between images with scene changes, the proposed method exploits DeepFlow with the extension of geometric constrains.
The estimated optical flow is exploited as input and its estimation error is modeled from training data.
The experimental results verified the effectiveness of the proposed methods.

The reason for the improvement of accuracy utilizing optical flow being small, especially for the TSUNAMI dataset, is that appearance changes owing to the difference of camera viewpoints are not large, and the ratio of the detection errors owing to the appearance change to the entire change region is small \footnote{It should be noted that the standard deviations of $F_1$-scores of CDNet and DOF-CDNet are 0.114 and 0.103 for TSUNAMI, 0.131 and 0.128 for GSV, respectively. The results indicate that optical flow information can make the scene change detection more stable.}.
Furthermore, the proposed method has errors for a wide variety of changes of scenes and camera viewpoints, because of the lack of the training data. 
Therefore, we plan to create a large-scale change detection dataset that contains a wide variety of camera viewpoint changes.
We will also improve and evaluate the proposed method under severer viewpoint conditions.

\bibliographystyle{ieeetr}
\small{
\bibliography{strings,refs}
}
\end{document}